\documentclass{clv3}
\usepackage{amsmath}
\usepackage{url}
\usepackage{tikz}
\usepackage{subfig}
\usepackage{float}
\usepackage{pgfplots}
\usepackage{stmaryrd}
\renewcommand\vec[1]{\overrightarrow{#1}}
\newcommand\cev[1]{\overleftarrow{#1}}
\usepackage{hyperref}
\usepackage{xcolor}
\usepackage{pgfplots}
\usetikzlibrary{pgfplots.groupplots}
\definecolor{darkblue}{rgb}{0, 0, 0.5}
\hypersetup{colorlinks=true,citecolor=darkblue, linkcolor=darkblue, urlcolor=darkblue}

\issue{xx}{xx}{2017}

\runningtitle{Learning an Executable Neural Semantic Parser}

\runningauthor{Cheng, Reddy, Saraswat and Lapata}

\begin{document}

\title{Learning an Executable Neural Semantic Parser}

\author{Jianpeng Cheng}
\affil{University of Edinburgh}

\author{Siva Reddy}
\affil{Stanford University}

\author{Vijay Saraswat}
\affil{IBM T.J. Watson Research}

\author{Mirella Lapata}
\affil{University of Edinburgh}

\maketitle

\begin{abstract}
  This paper describes a neural semantic parser that maps natural
  language utterances onto logical forms which can be executed against
  a task-specific environment, such as a knowledge base or a database,
  to produce a response.  The parser generates tree-structured logical
  forms with a transition-based approach which combines a generic
  tree-generation algorithm with domain-general grammar defined by
  the logical language.  The generation process is modeled by
  structured recurrent neural networks, which provide a rich encoding
  of the sentential context and generation history for making
  predictions.  To tackle mismatches between natural language and
  logical form tokens, various attention mechanisms are explored.
  Finally, we consider different training settings for the neural
  semantic parser, including  fully supervised training where
  annotated logical forms are given, weakly-supervised training where
  denotations are provided, and distant supervision where only
  unlabeled sentences and a knowledge base are available. Experiments
  across a wide range of datasets demonstrate the effectiveness of our
  parser.
\end{abstract}

\section{Introduction}
\label{sec:introduction}

An important task in artificial intelligence is to develop systems
that understand natural language and enable interactions between
computers and humans.  Semantic parsing has emerged as a key
technology towards achieving this goal.  Semantic parsers specify a
mapping between natural language utterances and machine-understandable
meaning representations, commonly known as \textit{logical forms}.  A
logical form can be executed
against a real-world \textit{environment}, such as a knowledge base,
to produce a response, often called a \textit{denotation}.
Table~\ref{tab:1} shows examples of natural language queries, their
corresponding logical forms, and denotations. The query \textsl{What
  is the longest river in Ohio?}  is represented by the logical
form \texttt{longest(and(type.river, location(Ohio)))}, which when
executed against a database of US geography returns the answer
\textsl{Ohio River}. In the second example, the logical form
\texttt{count(daughterOf(Barack Obama))} corresponds to the query
\textsl{How many daughters does Obama have?} and is executed against
the Freebase knowledge base to return the answer~\textsl{2}.

In recent years, semantic parsing has attracted a great deal of
attention due to its utility in a wide range of applications such as
question answering \cite{kwiatkowski2011lexical,liang2011learning},
relation extraction \cite{krishnamurthy2012weakly}, goal-oriented
dialog \cite{wen-EtAl:2015:EMNLP}, natural language interfaces
\cite{popescu-EtAl:2004:COLING}, robot control
\cite{Matuszek:etal:2012}, and interpreting instructions
\cite{chen2011learning,artzi-zettlemoyer:2013:TACL}.  

Early statistical semantic parsers
\cite{zelle1996learning,zettlemoyer:learning:2005,wong:learning:2006,kwiatkowksi2010inducing}
mostly requires training data  in the form of utterances paired with annotated logical
forms.  More recently, alternative forms of supervision have been
proposed to alleviate the annotation burden, e.g.,~training on
utterance-denotation pairs
\cite{clarke2010driving,liang2016learning,kwiatkowski2013scaling}, or
using distant supervision \cite{krishnamurthy2012weakly,cai2013large}.
Despite different supervision signals, training and inference procedures in
conventional semantic parsers rely largely on domain-specific
grammars and engineering. A CKY-style chart parsing algorithm is commonly
employed to parse a sentence in polynomial time. 


\begin{table}[t]
  \caption{\label{tab:1} Examples of questions, corresponding logical forms, and
    their answers.}
\begin{center}
\begin{tabular}{|l|} \hline
		{\em Environment:} A database of US geography \\
		{\em Utterance:} What is the longest river in Ohio?\\
		{\em Logical form:}  \texttt{longest(and(type.river, location(Ohio)))} \\
		{\em Denotation:} {Ohio River}
\\ \hline \hline
		{\em Environment:} Freebase \\
		{\em Utterance:} {How many daughters does Obama have?}\\
		{\em Logical form:}   \texttt{count(daughterOf(Barack Obama))}\\
		{\em Denotation:} {2} \\ \hline
	\end{tabular} 
\end{center}
\end{table}

The successful application of recurrent neural networks
\cite{bahdanau2014neural,sutskever2014sequence} to a variety of NLP
tasks has provided strong impetus to treat semantic parsing as a
sequence transduction problem where an utterance is mapped to a target
meaning representation in string format
\cite{dong2016language,jia2016data,kovcisky2016semantic}.   Neural
semantic parsers generate a sentence in linear time, while reducing the need for domain-specific assumptions,
grammar learning, and more generally extensive feature
engineering. But this modeling flexibility comes at a cost since it is
no longer possible to interpret how meaning composition is
performed, given that logical forms are structured objects like trees or graphs. Such knowledge plays a critical role in understanding
modeling limitations so as to build better semantic parsers. Moreover,
without any task-specific knowledge, the learning problem is fairly
unconstrained, both in terms of the possible derivations to consider
and in terms of the target output which can be syntactically invalid.

In this work we propose a neural semantic parsing framework which
combines recurrent neural networks and their ability to model
long-range dependencies with a transition system to generate
well-formed and meaningful logical forms.  The transition system
combines a generic tree-generation algorithm with a small set of
domain-general grammar pertaining to the logical language to
guarantee correctness. Our neural parser differs from conventional
semantic parsers in two respects. Firstly, it does not require
lexicon-level rules to specify the mapping between natural language
and logical form tokens.  Instead, the parser is designed to handle
cases where the lexicon is missing or incomplete thanks to a neural
attention layer, which encodes a soft mapping between natural language
and logical form tokens.  This modeling choice greatly reduces the
number of grammar rules used during inference to those only specifying
domain-general aspects.  Secondly, our parser is transition-based
rather than chart-based. Although chart-based inference has met with
popularity in conventional semantic parsers, it has difficulty in
leveraging sentence-level features since the dynamic programming
algorithm requires features defined over substructures.  In
comparison, our linear-time parser allows us to generate parse
structures incrementally conditioned on the entire sentence.

We perform several experiments in downstream question-answering tasks
and demonstrate the effectiveness of our approach across different
training scenarios. These include \emph{full supervision} with
questions paired with annotated logical forms using the
\textsc{GeoQuery} \cite{zettlemoyer:learning:2005} dataset, \emph{weak
  supervision} with question-answer pairs using the
\textsc{WebQuestions} \cite{berant2013semantic} and
\textsc{GraphQuestions} \cite{su2016generating} datasets and
\emph{distant supervision} without question-answer pairs, using the
\textsc{spades} \cite{bisk2016evaluating} dataset. Experimental
results show that our neural semantic parser is able to generate high
quality logical forms and answer real-world questions on a wide range
of domains.

The remainder of this article is structured as
follows. Section~\ref{sec:related-work} provides an overview of
related work. Section~\ref{sec:neur-semant-pars} introduces our neural
semantic parsing framework and discusses the various training
scenarios to which it can be applied. Our experiments are described in
Section~\ref{sec:experiments} together with detailed analysis of
system output. Discussion of future work concludes the paper in
Section~\ref{sec:disc-concl}.

\section{Related Work}
\label{sec:related-work}

The proposed framework has connections to several lines of research
including various formalisms for representing natural language
meaning, semantic parsing models, and the training regimes they
adopt. We review related work in these areas below.

\paragraph{Semantic Formalism} 

Logical forms have played an important role in semantic parsing
systems since their inception in the 1970s
\cite{Winograd:1972,woods1972lunar}. The literature is rife with
semantic formalisms which can be used to define logical
forms. Examples include lambda calculus \cite{Montague:1973} which has
been used by many semantic parsers
\cite{zettlemoyer:learning:2005,kwiatkowksi2010inducing,reddy2014large}
due to its expressiveness and flexibility to construct logical forms of
great complexity, 
Combinatory Categorial Grammar \cite{steedman2000syntactic},
dependency-based compositional semantics
\cite{liang2011learning}, frame semantics \cite{baker1998berkeley} and
abstract meaning representations \cite{banarescu2013abstract}.

In this work, we adopt a database querying language as the
semantic formalism, namely the functional query language (FunQL; \citet{zelle1995using}).
FunQL maps first-order logical forms into
function-argument structures, resulting in recursive, tree-structured, program representations.
Although it lacks expressive power, FunQL has a modeling advantage for downstream tasks,
 since it is more natural to describe the manipulation of a simple world as procedural programs.
This modeling advantage has been revealed in recent advances of neural programmings: recurrent neural networks 
have demonstrated great capability in inducing compositional programs \cite{reed2015neural,neelakantan2015neural,Cai2017Making}. For example, they
learn to perform grade-school additions, bubble sort and table comprehension in procedures. 
Finally, some recent work
\cite{yin2017syntactic,iyer2017learning,zhong2017seq2sql} 
uses other programming languages, such as the SQL as the semantic formalism.

\paragraph{Semantic Parsing Model}

The problem of learning to map utterances to meaning representations
has been studied extensively in the NLP community. Most data-driven
semantic parsers consist of three key components: a grammar, a
trainable model, and a parsing algorithm. The grammar defines the
space of derivations from sentences to logical forms, and the model
together with the parsing algorithm find the most likely
derivation. The model, which can take for example the form of an SVM \cite{kate2006using}, a structured
perceptron \cite{zettlemoyer2007online,lu2008generative,reddy2014large,reddy2016transforming} or a
log-linear model \cite{zettlemoyer:learning:2005,berant2013semantic},
scores the set of candidate derivations generated from the grammar.  During
inference, a chart-based parsing algorithm is commonly used to predict the most
likely semantic parse for a sentence.

With recent advances in neural networks and deep learning, 
there is a trend of reformulating semantic parsing as a machine translation problem.
The idea is not novel, since semantic parsing has been previously studied with statistical machine translation
approaches in both \citet{wong:learning:2006} and \citet{andreas_semantic_2013}.
However, the task setup is important to be revisited since recurrent neural networks 
have been shown to be extremely useful in context modeling and sequence generation \cite{bahdanau2014neural}. 
Following this direction, \citet{dong2016language} and \citet{jia2016data} develop neural semantic parsers which treat semantic parsing as 
a sequence to sequence learning problem. 
\citet{jia2016data}  also introduces a data augmentation approach which bootstraps a synchronous grammar from existing data and generates artificial examples as extra training data. 
Other related work extends the vanilla sequence to sequence model in various ways, 
such as
multi-task learning \cite{fan2017transfer},
parsing cross-domain queries \cite{herzig2017neural} and context-dependent queries \cite{suhr2018learning}, 
and applying the model to other formalisms such as AMR \cite{konstas2017neural} and SQL  \cite{zhong2017seq2sql}.

The fact that logical forms have a syntactic structure has motivated some of the recent work
on exploring structured neural decoders to generate tree or graph structures, 
and grammar constrained decoders to ensure the outputs are meaningful and executable. 
Related work includes \citet{yin2017syntactic} who generate abstract syntax trees for source code with a grammar constrained neural decoder. 
\citet{krishnamurthy2017neural} also introduce a neural semantic parser which decodes rules in a grammar to obtain well-typed logical forms.
\citet{Rabinovich-Stern-Klein:2017:AbstractSyntaxNetworks} propose
abstract syntax networks with a modular decoder, whose multiple submodels (one
per grammar construct) are composed to generate abstract syntax trees
in a top-down manner. 

Our work shares similar motivation: we generate tree-structured, syntactically valid logical
forms, however, following a transition-based generation approach
\cite{dyer2016recurrent,dyer2015transition}. Our semantic parser is a
generalization of the model presented in
\citet{cheng2017learning}. While they focus solely on top-down
generation using hard attention, the parser presented in this work
generates logical forms following either a top-down or bottom-up
generation order and introduces additional attention mechanisms
(i.e.,~soft and structured attention) for handling mismatches between
natural language and logical form tokens. We empirically compare
generation orders and attention variants, elaborate on model details,
and formalize how the neural parser can be effectively trained under
different types of supervision.

\paragraph{Training Regimes}

Various types of supervision have been explored to train semantic
parsers, ranging from full supervision with utterance-logical form pairs to
unsupervised semantic parsing without given utterances.  Early work of statistical semantic parsing has mostly
used annotated training data consisting of
utterances paired with logical forms
\cite{zelle1996learning,kate2006using,kate2005learning,wong:learning:2006,lu2008generative,kwiatkowksi2010inducing}.
Same applies to some of the recent work on neural semantic parsing \cite{dong2016language,jia2016data}.
This form of supervision is the most effective to train the parser, but is also expensive to obtain.
In order to write down a correct logical form, the annotator not only needs to have expertise in the semantic formalism,  
but also has to ensure the logical form matches the utterance semantics and contains no grammatical mistakes.
For this reason, fully supervised training applies more to small, close domain problems, such as querying the US geographical database \cite{zelle1996learning}.

Over the past few years, developments have been made to 
train semantic parsers with weak supervision from utterance-denotation pairs \cite{clarke2010driving,liang2011learning,berant2013semantic,kwiatkowski2013scaling,pasupat2015compositional}. 
The approach enables more efficient data collection, since denotations (such as answers to a question, responses to a system)
are much easier to obtain via crowd sourcing.
For this reason, semantic parsing can be scaled to handle large, complex and open domain problems.
Examples include the work that learn semantic parsers from question-answer pairs on Freebase \cite{liang2011learning,berant2013semantic,berant2014semantic,liang2017neural,cheng2017learning},
from system feedbacks \cite{clarke2010driving,chen2011learning,artzi-zettlemoyer:2013:TACL}, from abstract examples \cite{goldman2017weakly}, and from human feedbacks \cite{iyer2017learning} or statements \cite{artzi-zettlemoyer:2011:EMNLP}.

Some work seeks for more clever ways of gathering data and trains semantic parsers with even weaker supervision. 
In a class of distant supervision methods, the input is solely a knowledge base and a corpus of unlabeled sentences. 
Artificial training data is generated from the given resources. 
For example, \citet{cai2013large} generate utterance paired with logical forms. Their approach searches for sentences containing certain entity pairs,
and assume (with some pruning technique) the sentences express a certain relation from the KB. 
In \citet{krishnamurthy2012weakly} and \citet{Krishnamurthy2014Joint} whose authors work with the CCG formalism, an extra source of supervision is added. The
semantic parser is trained to produce parses that syntactically agree with dependency structures.
\citet{reddy2014large} generate utterance-denotation pairs by
masking entity mentions in declarative sentences from a large
corpus. A semantic parser is then trained to predict the denotations
corresponding to the masked entities.


%

\section{Neural Semantic Parsing Framework}
\label{sec:neur-semant-pars}

We present a neural-network based semantic parser that maps an
utterance into a logical form, which can be executed in the context of
a knowledge base to produce a response.  Compared to traditional
semantic parsers, our framework reduces the amount of manually
engineered features and domain-specific rules.  As semantic formalism,
we choose the functional query language (FunQL), which is recursive
and tree-structured (Section~\ref{lf}).  A transition-based tree
generation algorithm is then defined to generate FunQL logical forms
(Sections~\ref{gtga}--\ref{constraints}).  The process of generating
logical forms is modeled by recurrent neural networks---a powerful
tool for encoding the context of a sentence and the generation history
for making predictions (Section \ref{nn}).  We handle mismatches between natural language and knowledge base through
various attention mechanisms
(Section~\ref{sec:next-token-pred}). Finally, we explore
different training regimes (Section~\ref{supervision}, including a
fully supervised setting where each utterance is labeled with
annotated logical forms, a weakly supervised setting where
utterance-denotation pairs are available, and distant supervision
where only a collection of unlabeled sentences and a knowledge base is
given.

\subsection{FunQL Semantic Representation\label{lf}}
As mentioned earlier, we adopt FunQL as our semantic formalism. FunQL
is a variable free recursive meaning representation language which
maps simple first order logical forms to function-argument structures that
abstract away from variables and quantifiers \cite{kate2006using}.
The language is also closely related to lambda DCS
\cite{liang2013lambda}, which makes existential quantifiers implicit.
Lambda DCS is more compact in the sense that it can use variables in
rare cases to handle anaphora and build composite binary predicates.

The FunQL logical forms we define contain the following primitive
functional operators. They overlap with simple lambda DCS
\cite{berant2013semantic} but differ slightly in syntax to ease
recursive generation of logical forms. Let $l$ denote a logical form,
$\llbracket l \rrbracket$ represent its denotation, and $\mathcal{K}$
refers to a knowledge base.

\begin{itemize}
\item Unary base case: An \texttt{entity} $e$ (e.g.,~\texttt{Barack Obama}) is a unary logical form
  whose denotation is a singleton set containing that entity: 
  \begin{equation}
  \llbracket e \rrbracket = \{e\}
  \end{equation}
  
\item Binary base case: A \texttt{relation} $r$
  (e.g.,~\texttt{daughterOf})  is a binary logical form with
  denotation: 
\begin{equation}
\llbracket r \rrbracket = \{(e_1, e_2): (e_1, r, e_2) \in \mathcal{K}\}
\end{equation}

\item A relation $r$ can be applied to an entity $e_1$ (written as
  $r(e_1)$) and returns as denotation the unary satisfying the
  relation:
\begin{equation}
  \llbracket r(e_1) \rrbracket = \{e: (e_1, e) \in \llbracket r \rrbracket \}
\end{equation}  
   For example,
  the expression \texttt{daughterOf(Barack Obama)} corresponds to the question
  ``\textsl{Who are Barack Obama's daughters?}''. 

\item \texttt{count} returns the cardinality of the unary set $u$:  
\begin{equation}
\llbracket \texttt{count}(u) \rrbracket = \{ | \llbracket u \rrbracket | \}
\end{equation} 
For example, \texttt{count(daughterOf(Barack Obama))} represents the
question ``\textsl{How many daughters does Barack Obama have?}''.

\item \texttt{argmax} or \texttt{argmin} return a subset of the unary
  set $u$ whose specific relation $r$ is maximum or minimum:
\begin{equation}    
  \llbracket \texttt{argmax}(u, r) \rrbracket= \{ e: e \in u \cap \forall e' \in u, r(e) \geq r(e')   \}
  \end{equation} 
  For example, the expression \texttt{argmax(daughterOf(Barack Obama), age)}
  corresponds to the utterance ``\textsl{Who is Barack Obama's eldest
    daughter?}''.

\item \texttt{filter} returns a subset of the unary set $u$ where a comparative
  constraint ($=$, $!=$, $>$, $<$, $\geq$, $\leq$) acting on the relation $r$ is satisfied: 
\begin{equation}      
  \llbracket \texttt{filter}_{>} (u, r, v) \rrbracket = \{ e: e \in u \cap r(e) > v \}
 \end{equation}  
  For example, the query
  \texttt{filter$_{>}$ (daughterOf(Barack Obama), age, 5)} returns the daughters of Barack Obama who are older than five
  years.

\item \texttt{and} takes the intersection of two urinary sets $u_1$ and $u_2$:  
\begin{equation}   
\llbracket \texttt{and}(u_1, u_2) \rrbracket = \llbracket u_1 \rrbracket \cap \llbracket u_2 \rrbracket
\end{equation} 
while \texttt{or} takes their union:
  \begin{equation}  
  \llbracket \texttt{or}(u_1, u_2) \rrbracket = \llbracket u_1 \rrbracket \cup \llbracket u_2 \rrbracket
  \end{equation} 
  For example, the expression \texttt{and(daughterOf(Barack Obama),
    InfluentialTeensByYear(2014))} would correspond to the query
  ``\textsl{Which daughter of Barack Obama was named Most Influential
    Teens in the year 2014?}''.
\end{itemize}
The operators just defined give rise to compositional logical
forms (e.g.,~\texttt{count(and(daughterOf(Barack Obama),
  InfluentialTeensByYear(2014))}.

The reason for using FunQL in our framework lies in its recursive
nature which allows us to model the process of generating logical form
as a sequence of transition operations, which can be decoded by
powerful recurrent neural networks.  We next describe how our semantic formalism
is integrated with a transition-based tree-generation algorithm to
produce tree-structured logical forms.

\subsection{Tree Generation Algorithm \label{gtga}}
We introduce a generic tree generation algorithm which recursively
generates tree constituents with a set of transition operations.  The
key insight underlying our algorithm is to define a \textit{canonical}
traversal or generation order, which generates a tree as a \textit{transition sequence}. 
A transition sequence for a tree is a sequence of configuration-transition pairs
[$(c_0, t_0), (c_1, t_1), \cdots, (c_m, t_m)$]. 
In this work, we consider two commonly
used generation orders, namely top-down pre-order and bottom-up
post-order. 

The \textbf{top-down} system is specified by the tuple
$c=(\sum, \pi, \sigma, N, P)$ where $\sum$ is a stack used to store
partially complete tree fragments, $\pi$ is non-terminal token to be
generated, $\sigma$ is the terminal token to be generated, $N$ is a
stack of open non-terminals, and $P$ is a function indexing the
position of a non-terminal pointer. The pointer indicates where
subsequent children nodes should be attached (e.g.,~$P(X)$ means that
the pointer is pointing to the non-terminal $X$).  We take the
parser's initial configuration to be
$c_0 = ([], TOP, \varepsilon, [], \bot)$, where $TOP$ stands for the
root node of the tree, $\varepsilon$ represents an empty string, and
$\bot$ represents an unspecified function. The top-down system employs
 three transition operations  defined in Table~\ref{td_formal}:

\begin {table}[t]
\begin{center}
	\begin{tabular}{l l }
		\hline
		\multicolumn{2}{c}{\textbf{Top-down Transitions}} \\\hline
		\texttt{NT(X)}  &  $([\sigma | \texttt{X}'], \texttt{X}, \varepsilon,  [\beta | \texttt{X}'], P(\texttt{X}')) \Rightarrow  ([\sigma | \texttt{X}', \texttt{X}], \varepsilon, \varepsilon, [\beta | \texttt{X}', \texttt{X}], P(\texttt{X})) $  \\
		\texttt{TER(x)}  & $([\sigma | \texttt{X}'], \varepsilon, \texttt{x}, [\beta | \texttt{X}'], \texttt{P(X')}) \Rightarrow  ( [\sigma | \texttt{X}', \texttt{x}], \varepsilon, \varepsilon , [\beta | \texttt{X}', \texttt{x}], P(\texttt{X}'))  $  \\
		\texttt{RED} & $([\sigma | \texttt{X}', \texttt{X}, \texttt{x}], \varepsilon, \varepsilon, [\beta | \texttt{X}', \texttt{X}], P(\texttt{X})) \Rightarrow  ([\sigma | \texttt{X}', \texttt{X}(\texttt{x})], \varepsilon, \varepsilon, [\beta | \texttt{X}'], P(\texttt{X}')) $ \\
		\hline
\multicolumn{2}{c}{} \\\hline
\multicolumn{2}{c}{\textbf{Bottom-up Transitions}} \\\hline
		\texttt{TER(x)}  & $(\sigma , \varepsilon, \texttt{x}) \Rightarrow  ([\sigma | \texttt{x}], \varepsilon, \varepsilon)   $  \\
		\texttt{NT-RED(X)} & $([\sigma | \texttt{x}], \texttt{X}, \varepsilon) \Rightarrow  ([\sigma | \texttt{X}(\texttt{x})], \varepsilon, \varepsilon) $ \\
		\hline
	\end{tabular}
\end{center}
\caption{Transitions for top-down and bottom-up generation system. Stack~$\sum$
  is represented as a list with its head to the right (with tail
  $\sigma$), same for stack~$N$ (with tail
  $\beta$). \label{td_formal}} 
\end{table}

\begin{itemize}
\item \texttt{NT(X)} creates a new subtree non-terminal node denoted
  by \texttt{X}.  The non-terminal \texttt{X} is pushed on top of the
  stack and written as \texttt{X(} while subsequent tree nodes are
  generated as children underneath~\texttt{X}.
\item \texttt{TER(x)} creates a new child
  node denoted by \texttt{x}. The terminal \texttt{x} is pushed on top
  of the stack, written as \texttt{x}.
\item \texttt{RED} is the reduce operation which indicates that the
  current subtree being generated is complete. The non-terminal root
  of the current subtree is closed and subsequent children nodes will
  be attached to the predecessor open non-terminal.  Stack-wise,
  \texttt{RED} recursively pops children (which can be either
  terminals or completed subtrees) on top until an open non-terminal
  is encountered.  The non-terminal is popped as well, after which a
  completed subtree is pushed back to the stack as a single closed
  constituent, written for example as \texttt{X1}(\texttt{X2},
  \texttt{X3}).
\end{itemize}

We define the \textbf{bottom-up} system by tuple
$c=(\sum, \pi, \sigma)$ where $\sum$ is a stack used to store
partially complete tree fragments, $\pi$ is the token non-terminal to
be generated, and $\sigma$ is the token terminal to be generated.  We
take the initial parser configuration to be
$c_0 = ([], x_l, \varepsilon)$, where $x_l$ stands for the leftmost
terminal node of the tree, and $\varepsilon$ represents an empty
string.  The bottom-up generation uses two transition operations
defined in Table~\ref{td_formal}:

\begin{itemize}
\item \texttt{TER(x)} creates a new terminal node denoted by
  \texttt{x}. The terminal \texttt{x} is pushed on top of the stack,
  written as \texttt{x}.
\item \texttt{NT-RED(X)} builds a new subtree by attaching a parent
  node (denoted by \texttt{X}) to children nodes on top of the stack.
  The children nodes can be either terminals or smaller subtrees.
  Similarly to \texttt{RED} in the top-down case, children nodes are
  first popped from the stack, and subsequently combined with the
  parent \texttt{X} to form a subtree.  The subtree is pushed back to
  the stack as a single constituent, written for example as
  \texttt{X1}(\texttt{X2}, \texttt{X3}). A challenge with
  \texttt{NT-RED(X)} is to decide how many children should be popped
  and included in the new subtree. In this work, the number of
  children is dictated by the number of arguments expected by
  \texttt{X} which is in turn constrained by the logical language.
  For example, from the FunQL grammar it is clear that \texttt{count}
  takes one argument and \texttt{argmax} takes two.  The language we
  use does not contain non-terminal functions with a variable number of
  arguments.
\end{itemize}

Top-down traversal is defined by three generic operations, while
bottom-up order applies two operations only (since it combines reduce
with non-terminal generation). However, the operation predictions
required are the same for the two systems.  The reason is that the
reduce operation in the top-down system is deterministic when the
FunQL grammar is used as a constraint (we return to this point in
Section~\ref{constraints}).

\subsection{Generating Tree-structured Logical Forms}
To generate tree-structured logical forms, we integrate the generic
tree generation operations described above with FunQL, whose grammar
determines the space of allowed terminal and non-terminal symbols:
\begin{itemize}
\item \texttt{NT(X)} includes an operation that generates
  relations \texttt{NT(relation)}, and other domain-general operators in FunQL:
  \texttt{NT(and)}, \texttt{NT(or)}, \texttt{NT(count)},
  \texttt{NT(argmax)}, \texttt{NT(argmin)} and \texttt{NT(filter)}.
  Note that \texttt{NT(relation)} creates a placeholder for a relation, which is subsequently generated.

\item \texttt{TER(X)} includes two operations:
  \texttt{TER(relation)} for generating relations and
  \texttt{TER(entity)} for generating entities. Both operations create a placeholder for a relation or an entity, which is subsequently generated.

\item \texttt{NT-RED(X)} includes \texttt{NT-RED(relation)},
  \texttt{NT-RED(and)}, \texttt{NT-RED(or)}, \texttt{NT-RED(count)},
  \texttt{NT-RED(argmax)}, \texttt{NT-RED(argmin)} and
  \texttt{NT-RED(filter)}. Again, \texttt{NT-RED(relation)} creates a
  placeholder for a relation, which is subsequently generated.
\end{itemize}

Table~\ref{tdeg} illustrates the sequence of operations employed by
our parser in order to generate the logical form
\texttt{count(and(daughterOf(Barack Obama),
  InfluentialTeensByYear(2014))} top-down.  Table~\ref{bueg} shows how
the same logical form is generated bottom-up. Note that the examples
are simplified for illustration purposes; the logical form is
generated \emph{conditioned} on an input utterance, such as
``\textsl{How many daughters of Barack Obama were named Most
  Influential Teens in the year 2014?}''.

\begin {table}[t]
  \caption{Top-down generation of the logical form
    \texttt{count(and(daughterOf(Barack Obama),
      InfluentialTeensByYear(2014))}. Elements on the stack are separated by $||$ and the top of the stack is on the right.} 
\label{tdeg}
\begin{center}
	\scriptsize
	\begin{tabular}{lcl}
		\hline
		Operation & Logical form token & \multicolumn{1}{c}{Stack}\\ \hline 
		\texttt{NT(count)} & \texttt{count} & \texttt{count}( \\
		\texttt{NT(and)} & \texttt{and} & \texttt{count}( $||$ \texttt{and}( \\
		\texttt{NT(relation)} & \texttt{daughterOf} & \texttt{count}( $||$ \texttt{and}( $||$ \texttt{daughterOf} \\
		\texttt{TER(entity)} & \texttt{Barack Obama} & \texttt{count}( $||$ \texttt{and}( $||$ \texttt{daughterOf}( $||$ \texttt{Barack Obama} \\
		\texttt{RED} & & \texttt{count}( $||$ \texttt{and}( $||$ \texttt{daughterOf}(\texttt{Barack Obama}) \\
		\texttt{NT(relation)} & \texttt{InfluentialTeensByYear} & \texttt{count}( $||$ \texttt{and}( $||$ \texttt{daughterOf}(\texttt{Barack Obama}) $||$ \texttt{InfluentialTeensByYear}(  \\
		\texttt{TER(entity)} & \texttt{2014} & \texttt{count}( $||$ \texttt{and}( $||$ \texttt{daughterOf}(\texttt{Barack Obama}) $||$ \texttt{InfluentialTeensByYear}( $||$ \texttt{2014}  \\
		\texttt{RED} & & \texttt{count}( $||$ \texttt{and}( $||$ \texttt{daughterOf}(\texttt{Barack Obama}) $||$ \texttt{InfluentialTeensByYear}(\texttt{2014})  \\
		\texttt{RED} & & \texttt{count}( $||$ \texttt{and}(\texttt{daughterOf}(\texttt{Barack Obama}), \texttt{InfluentialTeensByYear}(\texttt{2014}))  \\
		\texttt{RED} & & \texttt{count}(\texttt{and}(\texttt{daughterOf}(\texttt{Barack Obama}), \texttt{InfluentialTeensByYear}(\texttt{2014})))  \\
		\hline
	\end{tabular}
\end{center}
\end{table}

\begin {table}[t]
  \caption{Bottom-up generation of the logical form
    \texttt{count(and(daughterOf(Barack Obama),
      InfluentialTeensByYear(2014))}. Elements on the stack are
    separated by $||$ and the top of the stack is on the right.} 
\label{bueg}
\begin{center}
	\scriptsize
	\begin{tabular}{lcl}
		\hline
		Operation & Logical form token & \multicolumn{1}{c}{Stack} \\ \hline 
		\texttt{TER(entity)} & \texttt{Barack Obama} & \texttt{Barack Obama} \\
		\texttt{NT-RED(relation)} & \texttt{daughterOf} & \texttt{daughterOf}(\texttt{Barack Obama}) \\
		\texttt{TER(entity)} & \texttt{2014} & \texttt{daughterOf}(\texttt{Barack Obama}) $||$ \texttt{2014} \\
		\texttt{NT-RED(relation)} & \texttt{InfluentialTeensByYear} & \texttt{daughterOf}(\texttt{Barack Obama}) $||$ \texttt{InfluentialTeensByYear}(\texttt{2014})  \\
		\texttt{NT-RED(and)} & \texttt{and} & \texttt{and}(\texttt{daughterOf}(\texttt{Barack Obama}), \texttt{InfluentialTeensByYear}(\texttt{2014}))  \\
		\texttt{NT-RED(count)} & \texttt{count} & \texttt{count}(\texttt{and}(\texttt{daughterOf}(\texttt{Barack Obama}), \texttt{InfluentialTeensByYear}(\texttt{2014})))  \\
		\hline
	\end{tabular}
\end{center}
\end{table}

\subsection{Constraints \label{constraints}}
A challenge in neural semantic parsing lies in generating well-formed
and meaningful logical forms.  To this end, we incorporate two types
of constraints in our system.  The first ones are structural
constraints to ensure that the outputs are syntactically valid logical
forms. For the top-down system these constraints include:
\begin{itemize}
	\item The first operation must be \texttt{NT};
	\item \texttt{RED} cannot directly follow \texttt{NT};
	\item The maximum number of open non-terminal symbols allowed
          on the stack is~10. \texttt{NT} is disabled when the maximum
          number is reached;  
	\item The maximum number of (open and closed) non-terminal
          symbols allowed on the stack is~10. \texttt{NT} is disabled
          when the maximum number is reached.
\end{itemize}
Tree  constraints for the bottom-up system are:
\begin{itemize}
	\item The first operation must be \texttt{TER};
	\item The maximum number of consecutive \texttt{TER}s allowed
          is~5; 
	\item The maximum number of terminal symbols allowed on the
          stack is the number of words in the sentence. \texttt{TER}
          is disallowed when the maximum number is reached. 
\end{itemize}
The second type of constraints relate to the FunQL-grammar itself, ensuring that
the generated logical forms are meaningful for execution:
\begin{itemize}
\item The type of argument expected by each non-terminal symbol must
  follow the FunQL grammar;
	\item The number of arguments expected by each non-terminal
          symbol must follow the FunQL grammar; 
	\item When the expected number of arguments for a non-terminal
          symbol is reached, a \texttt{RED} operation must be called
          for the top-down system; for the bottom-up system this
          constrain is built within the \texttt{NT-RED} operation,
          since it reduces the expected number of arguments based on a
          specific non-terminal symbol.
\end{itemize}

\subsection{Neural Network Realizer \label{nn}}
We model the above logical form generation algorithm with a structured
neural network which encodes the utterance and the generation history,
and then predicts a sequence of transition operations as well as
logical form tokens based on the encoded information. In the
following, we present details for each component in the network.

\paragraph{Utterance Encoding} An utterance~$x$ is encoded with a
bidirectional LSTM architecture \cite{hochreiter1997long}.  A
bidirectional LSTM is comprised of a forward LSTM and a backward LSTM.
The forward LSTM processes a variable-length sequence
$x=(x_1, x_2, \cdots, x_n)$ by incrementally adding new content into a
single memory slot, with gates controlling the extent to which new
content should be memorized, old content should be erased, and current
content should be exposed. At time step~$t$, the memory~$\vec{c_t}$
and the hidden state~$\vec{h_t}$ are updated with the following
equations:
\begin{equation}
	\begin{bmatrix}
		i_t\\ f_t\\ o_t\\ \hat{c}_t
	\end{bmatrix} =
	\begin{bmatrix} \sigma\\ \sigma\\ \sigma\\ \tanh
	\end{bmatrix} W\cdot [\vec{h_{t-1}}, \, x_t]
	\label{beginlstm}
\end{equation}
\begin{equation} \vec{c_t} = f_t \odot \vec{c_{t-1}} +
	i_t \odot \hat{c}_t
\end{equation}
\begin{equation} \vec{h_t} = o_t \odot \tanh(\vec{c_t})
	\label{endlstm}
\end{equation} 
where $i$, $f$, and $o$ are gate activations; $W$ denotes the weight matrix. For simplicity, we denote the recurrent computation of the forward LSTM as:
\begin{equation} \vec{h_t} = \vec{\textnormal{LSTM}} (x_t, \vec{h_{t-1}})
	\label{lstm1}
\end{equation} 
After encoding, a list of token representations $[\vec{h_1},
\vec{h_2}, \cdots, \vec{h_n}]$ within the forward context is obtained.
Similarly, the backward LSTM computes a list of token representations 
$[\cev{h_1}, \cev{h_2}, \cdots, \cev{h_n}]$ within the backward context as:
\begin{equation} \cev{h_t} = \cev{\textnormal{LSTM}} (x_t, \cev{h_{t+1}})
	\label{lstm2}
\end{equation} 

Finally, each input token $x_i$ is represented by the concatenation of
its forward and backward LSTM state vectors, denoted by $h_i =
\vec{h_i} : \cev{h_i}$.  The list storing token vectors for the entire
utterance $x$ can be considered as a buffer, in analogy to syntactic
parsing.  A notable difference is that tokens in the buffer will not
be removed since its alignment to logical form tokens is not
pre-determined in the general semantic parsing scenario.  We denote
the buffer $b$ as $b = [h_1, \cdots, h_k]$, where $k$ denotes the
length of the utterance.

\paragraph{Generation History Encoding} The generation history, aka
partially completed subtrees, is encoded with a variant of stack-LSTM
\cite{dyer2015transition}.  Such an encoder captures not only
previously generated tree tokens but also tree structures.  We first
discuss the stack-based LSTM in the top-down transition system and
then present modifications to account for the bottom-up system.

\begin{figure}[t]
	\centering
	\includegraphics[width=1\textwidth]{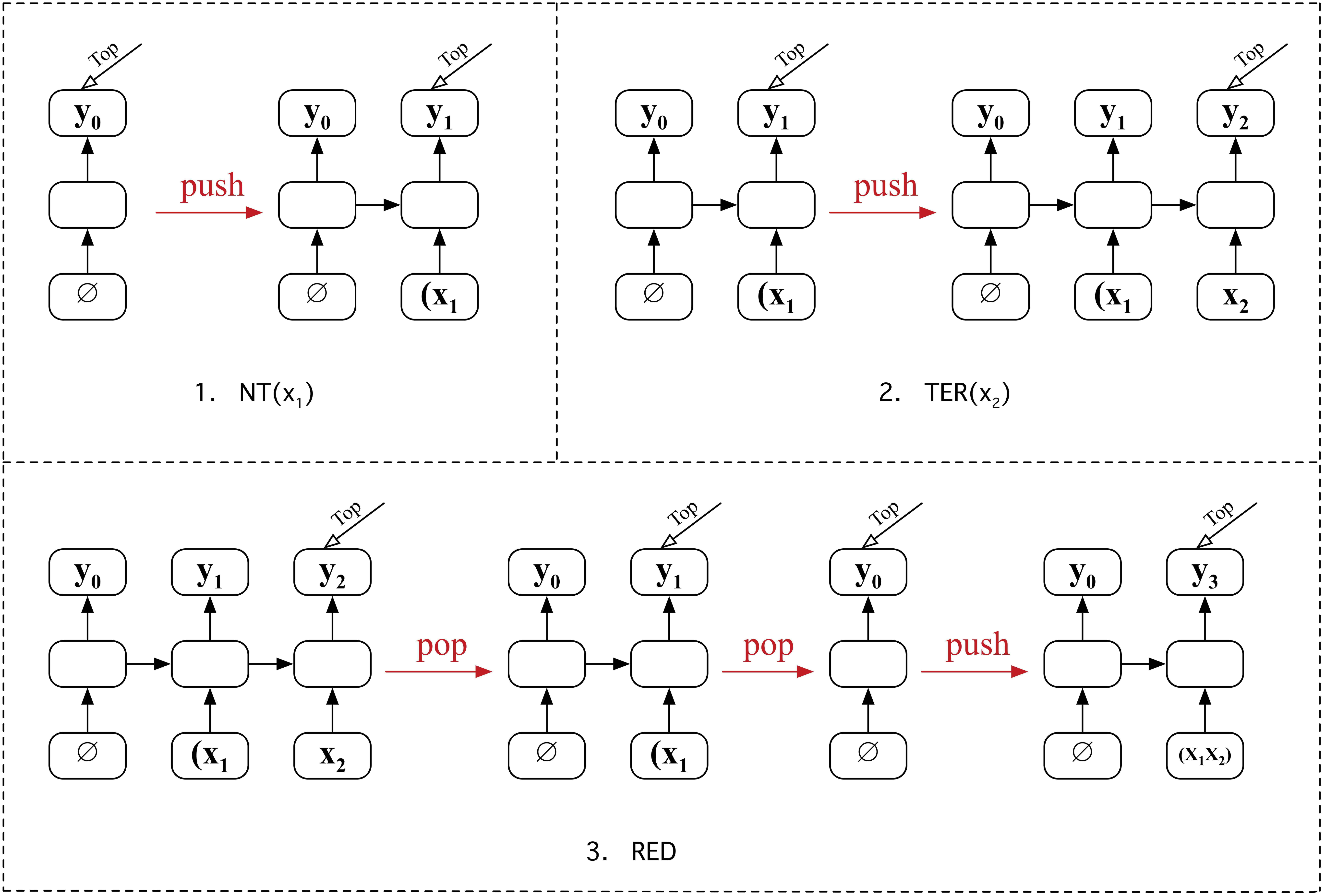}
	\caption{A stack-LSTM extends a standard LSTM with the
          addition of a stack pointer (shown as Top in the
          figure). The example shows how the configuration of the
          stack changes when the operations \texttt{NT}, \texttt{TER},
          and \texttt{RED} are applied in sequence. The initial stack
          is presumed empty for illustration purposes.  We only show
          how the stack-LSTM updates its states, not how subsequent
          predictions are made which depend not only on the hidden
          state of the stack-LSTM, but also on the natural language
          utterance.}
	\label{slstm}
\end{figure}

In \textbf{top-down} transitions, operations \texttt{NT} and \texttt{TER}
 change the stack-LSTM representation $s_t$ as in a vanilla LSTM as:
\begin{equation}
	s_t = \textnormal{LSTM} (y_t, s_{t-1})
\end{equation}
where~$y_t$ denotes the newly generated non-terminal or terminal
token. A \texttt{RED} operation recursively pops the stack-LSTM states
as well as corresponding tree tokens on the output stack. The popping
stops when a non-terminal state is reached and popped, after which the
stack-LSTM reaches an intermediate state $s_{t-1:t}$.\footnote{We use $s_{t-1:t}$ to denote the intermediate transit state from time step $t-1$ to $t$, after terminal tokens are popped from the stack; $s_t$ denotes the final LSTM state after the subtree representation is pushed back to the stack (as explained in the following).} The
representation of the completed subtree~$u$ is then computed as:
\begin{equation}
	u = W_u \cdot [p_u : c_u]
\end{equation}
where $p_u$ denotes the parent (non-terminal) embedding of the
subtree, $c_u$ denotes the average of the children (terminal or
completed subtree) embeddings, and $W_u$ denotes the weight matrix.  
Note that $c_u$ can also be computed with more advanced method such as a recurrent neural network \cite{kuncoro2016recurrent}. 
Finally, the subtree embedding~$u$ serves as the input to the
LSTM and updates $s_{t-1:t}$ to $s_t$ as:
\begin{equation}
	s_t = \textnormal{LSTM} (u, s_{t-1:t})
	\label{reduce:1}
\end{equation}
Figure \ref{slstm} provides a graphical view on how the three operations change the configuration of a stack-LSTM. 

In comparison, the \textbf{bottom-up} transition system uses the same
\texttt{TER} operation to update the stack-LSTM representation~$s_t$ when a terminal $y_t$ is newly generated:
\begin{equation}
	s_t = \textnormal{LSTM} (y_t, s_{t-1})
\end{equation}

Differently, the effects of \texttt{NT} and \texttt{RED} are merged into a \texttt{NT-RED(X)} operation.
When \texttt{NT-RED(X)} is invoked, a non-terminal $y_t$ is first predicted 
and then the stack-LSTM starts popping its states on the stack.
The number of pops is decided by the amount of argument expected by $y_t$. 
After that, a subtree can be obtained by combining the non-terminal $y_t$ and the newly popped terminal tokens, while the
stack-LSTM reaches an intermediate state $s_{t-1:t}$. Similar to the top-down system, we compute the
representation of the newly combined subtree~$u$ as:
\begin{equation}
	u = W_u \cdot [p_u : c_u]
\end{equation}
where $p_u$ denotes the parent (non-terminal) embedding of the
subtree, $c_u$ denotes the average of the children (terminal or
completed subtree) embeddings, and $W_u$ denotes the weight matrix.  Finally, the subtree embedding $u$~serves as the input to the
LSTM and updates $s_{t-1:t}$ to $s_t$ as:
\begin{equation}
	s_t = \textnormal{LSTM} (u, s_{t-1:t})
	\label{reduce:2}
\end{equation}
The key difference here is that a non-terminal
tree token is never pushed alone to update the stack-LSTM, but rather
as part of a completed subtree that does the update.

\paragraph{Making Predictions} 
Given encodings of the utterance and generation history, our model
makes two types of predictions pertaining to transition operations and
logical form tokens (see Tables~\ref{tdeg}, \ref{bueg}).  First, at
every time step, the next transition operation~$o_{t+1}$ is predicted
based on utterance encoding~$b$ and generation history~$s_t$:
\begin{equation}
	o_{t+1}  \sim f(b, s_t)
\end{equation}
where $f$ is a neural network that computes the parameters of a
multinomial distribution over the action space which is restricted by
the constraints discussed in Section~\ref{constraints}. 

Next, the logical form token underlying each generation operation must
be emitted. When the generation operation contains one of the
domain-general non-terminals \texttt{count}, \texttt{argmax},
\texttt{argmin}, \texttt{and}, \texttt{or}, and \texttt{filter}
(e.g.,~\texttt{NT(count)}), the logical form token is the
corresponding non-terminal (e.g.,~\texttt{count}).  When the
generation operation involves one of the placeholders for
\texttt{entity} or \texttt{relation} (e.g.,~\texttt{NT(relation)}, \texttt{NT-RED(relation)}, \texttt{TER(relation)} and \texttt{TER(entity)}), a
domain-specific logical form token~$y_{t+1}$ (i.e.,~an entity or a relation) is
predicted in a fashion similar to  action prediction:
\begin{equation}
	y_{t+1} \sim g(b, s_t)
\end{equation}
where $g$ is a neural network that computes the parameters of a
multinomial distribution over the token space.

A remaining challenge lies in designing predictive functions~$f$ (for
the next action) and~$g$ (for the next logical form token) in the
context of semantic parsing. We explore various attention mechanisms
which we discuss in the next sections.


\subsection{Next Action Prediction}
\label{sec:next-acti-pred}
This section explains how we model function~$f$ for predicting the
next action.  We draw inspiration from previous work on
transition-based syntactic parsing and compute a feature vector
representing the current state of the generation system
\cite{dyer2016recurrent}. This feature vector typically leverages the
buffer which stores unprocessed tokens in the utterance and the stack
which stores tokens in the partially completed parse tree. A major
difference in our semantic parsing context is that the buffer
configuration does not change deterministically with respect to the
stack since the alignment between natural language tokens and
logical-form tokens is not explicitly specified. This gives rise to
the challenge of extracting features representing the buffer at
different time steps.  To this end, we compute at each time step~$t$ a
single adaptive representation of the buffer~$\bar{b}_t$ with a soft
attention mechanism:
\begin{equation}
	u_t^i  = V \tanh (W_b b_i + W_s s_t) 
\end{equation}
\begin{equation}
	\alpha_t^i  =  \textnormal{softmax} (u_t^i )
\end{equation}
\begin{equation}
	\bar{b}_t  = \sum_i  \alpha_t^i  b_i 
	\label{average1}
\end{equation}
where $W_b$ and $W_s$ are weight matrices and $V$ is a weight vector.
We then combine the representation of the buffer and the stack with a
feed-forward neural network (Equation~(\ref{softmax})) to yield a
feature vector for the generation system. Finally,
$\textnormal{softmax}$ is taken to obtain the parameters of the
multinomial distribution over actions:
\begin{equation}
	a_{t+1}  \sim \textnormal{softmax} (  W_{oa} \tanh( W_f [\bar{b}_t, s_t] )  )
	\label{softmax}
\end{equation}
where $W_{oa}$ and $W_f$ are weight matrices.

\subsection{Next Token Prediction}
\label{sec:next-token-pred}
This section presents various functions~$g$ for predicting the next
logical form token (i.e.,~a specific \texttt{entity} or
\texttt{relation}).  A hurdle in semantic parsing concerns handling
mismatches between natural language and logical tokens
in the target knowledge base.  For example, both
utterances ``\textit{Where did X graduate from}'' and ``\textit{Where
  did X get his PhD}'' would trigger the same predicate
\texttt{education} in Freebase.  Traditional semantic parsers map
utterances directly to domain-specific logical forms relying
exclusively on a set of lexicons either predefined or learned for the
target domain with only limited coverage. Recent approaches alleviate
this issue by firstly mapping the utterance to a domain-general
logical form which aims to capture language-specific semantic aspects,
after which ontology matching is performed to handle mismatches
\cite{kwiatkowski2013scaling,reddy2014large,reddy2016transforming}.
Beyond efficiency considerations, it remains unclear which
domain-general representation is best suited to domain-specific
semantic parsing.

Neural networks provide an alternative solution: the matching between
natural language and domain-specific predicates is accomplished via an
attention layer, which encodes a context-sensitive probabilistic
lexicon.  This is analogous to the application of the attention
mechanism in machine translation \cite{bahdanau2014neural}, which is
used as an alternative to conventional phrase tables.  In this work,
we consider a practical domain-specific semantic parsing scenario
where we are given no lexicon.  We first introduce the basic form of
attention used to predict logical form tokens and then discuss various
extensions as shown in Figure~\ref{attvis}.

\paragraph{Soft Attention}
In the case where no lexicon is provided, we use a soft attention
layer similar to action prediction. The parameters of the soft
attention layer prior to softmax are shared with those used in action
prediction:
\begin{equation}
	u_t^i  = V \tanh (W_b b_i + W_s s_t) 
	\label{uti}
\end{equation}
\begin{equation}
	\alpha_t^i  =  \textnormal{softmax} (u_t^i )
\end{equation}
\begin{equation}
	\bar{b}_t  = \sum_i  \alpha_t^i  b_i 
	\label{average2}
\end{equation}
\begin{equation}
	y_{t+1}  \sim \textnormal{softmax} (  W_{oy} \tanh( W_f [\bar{b}_t, s_t] )  )
\end{equation}
which outputs the parameters of the multinomial distribution over
logical form tokens (either predicates or entities). When dealing with extremely
large knowledge bases, the output
space can be pruned and restricted with an entity linking procedure. This method
requires us to identity potential entity candidates in the sentence,
and then generate only entities belonging to this subset and the
relations linking them.

\paragraph{Structured Soft Attention} We also explored a structured
attention layer \cite{kim2017structured,liu2017learning} to encourage
the model to attend to contiguous natural language phrases when
generating a logical token, while still being differentiable.

The structured attention layer we adopt is a linear-chain conditional random
field (CRF; \citet{lafferty2001conditional}.  Assume that at time step
$t$ each token in the buffer (e.g., the $i$th token) is assigned an
attention label $A^i_t \in \{0,1\}$.  The CRF defines~$p(A_t)$, the
probability of the sequence of attention labels at time step~$t$ as:
\begin{equation}
	p(A_t)= \frac{\exp \sum_i  W_f \cdot \psi(A_t^{i-1}, A_t^i, b_i, s_t) }{ \sum_{A_t^1,\cdots, A_t^n}  \exp \sum_i  W_f \cdot \psi(A_t^{i-1}, A_t^i, b_i, s_t) }
\end{equation}
where $\sum_i$ sums over all tokens and $\sum_{A_t^1,\cdots, A_t^n}$ sums over all possible sequences of attention labels.
$W_f$ is a weight vector and $\psi(A_t^{i-1}, A_t^i, b_i, s_t) $
a feature vector. 
In this work the feature vector is defined with
three dimensions: the {state feature} for each token:
\begin{equation}
	u_t^i \cdot a_t^i 
\end{equation}
where $u_t^i$ is the token-specific attention score computed in
Equation~\eqref{uti}; the  {transition feature}:
\begin{equation}
	A_{t}^{i-1}  \cdot A_{t}^i 
\end{equation}
and the context-dependent {transition feature}
\begin{equation}
	u_t^i \cdot A_t^{i-1} \cdot A_{t}^i 
\end{equation}

\begin{figure}[t]
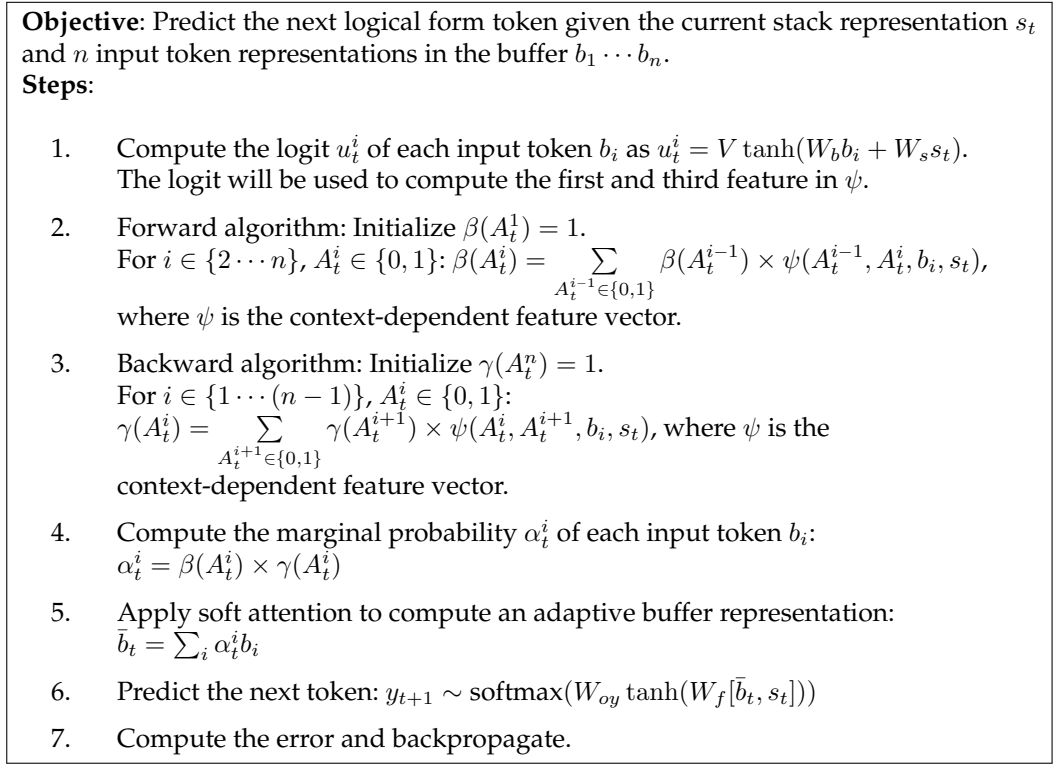

	\centering
	\fbox{
		\begin{minipage}{1\textwidth}
			\textbf{Objective}: Predict the next logical form token given the current stack representation $s_t$ and $n$~input token representations in the buffer $b_1\cdots b_n$. \\
			\textbf{Steps}: 
			\begin{enumerate}
				\item Compute the logit $u_t^i$ of each input token $b_i$ as $u_t^i  = V \tanh (W_b b_i + W_s s_t)$. The logit will be used to compute the first and third feature in $\psi$.
				\item Forward algorithm: Initialize $\beta(A_{t}^1) = 1$. \\ For $i \in \{2 \cdots n\}$, $A_{t}^i \in \{0, 1\}$: $\beta(A_{t}^i) = \sum\limits_{A_{t}^{i-1} \in \{0, 1\}} \beta(A_{t}^{i-1}) \times \psi(A_{t}^{i-1}, A_{t}^i, b_i, s_t)$, where $\psi$ is the context-dependent feature vector.
				\item Backward algorithm: Initialize $\gamma(A_t^n) = 1$. \\ For $i \in \{1 \cdots (n-1)\}$, $A_{t}^i \in \{0, 1\}$: $\gamma(A_{t}^i) = \sum\limits_{A_{t}^{i+1} \in \{0, 1\}} \gamma(A_{t}^{i+1}) \times \psi(A_{t}^i, A_{t}^{i+1}, b_i, s_t)$, where $\psi$ is the context-dependent feature vector.
				\item Compute the marginal probability $\alpha_t^i$ of each input token $b_i$: $\alpha_t^i = \beta(A_t^i) \times \gamma(A_t^i)$
				\item Apply soft attention to compute an adaptive buffer representation: $ \bar{b}_t  = \sum_i  \alpha_t^i  b_i $
				\item Predict the next token: $y_{t+1}  \sim \textnormal{softmax} (  W_{oy} \tanh( W_f [\bar{b}_t, s_t] )  )$
				\item Compute the error and backpropagate.
			\end{enumerate}
		\end{minipage}
	}
	\caption{The structured attention model for token prediction.}
	\label{stratt}
\end{figure}

The marginal probability $p(A_{t}^i = 1)$ of each token being selected
 is computed with the forward-backward message
passing algorithm \cite{lafferty2001conditional}. The procedure is
shown in Figure~\ref{stratt}.  To compare with standard soft
attention, we denote this procedure as:
\begin{equation}
	\alpha_t^i  =  \textnormal{forward-backward} (u_t^i )
\end{equation}

The marginal probabilities are used as in standard soft attention to
compute an adaptive buffer representation:
\begin{equation}
	\bar{b}_t  = \sum_i  \alpha_t^i  b_i 
	\label{average3}
\end{equation}
which is then used to compute a distribution of output logical form tokens:
\begin{equation}
	y_{t+1}  \sim \textnormal{softmax} (  W_{oy} \tanh( W_f [\bar{b}_t, s_t] )  )
\end{equation}

The structured attention layer is soft and fully differentiable and
allows us to model attention over phrases since the forward-backward
algorithm implicitly sums over an exponentially-sized set of
substructures through dynamic programming.

\paragraph{Hard Attention}
Soft attention learns a complete mapping between natural language and
logical tokens with a differentiable neural layer.  At every time
step, every natural language token in the utterance is assigned the
probability of triggering every logical predicate.  This offers little
in the way of interpretability. In order to render the inner workings
of the model more transparent we explore the use of a hard attention
mechanism as a means of rationalizing neural predictions.

At each time step, hard attention samples from the attention
probability a single natural language token~$x_t$:
\begin{equation}
	u_t^i  = V \tanh (W_b b_i + W_s s_t) 
\end{equation}
\begin{equation}
	x_t \sim \textnormal{softmax} (u_t^i )
\end{equation}
The representation of $x_t$ denoted by $b_t$ is then used to predict the logical token $y_t$:
\begin{equation}
	y_{t+1}  \sim \textnormal{softmax} ( W_{oy} \tanh( W_f [b_t, s_t] )  )
\end{equation}

Hard attention is nevertheless optimization-wise challenging; it
requires sampling symbols (aka non-differentiable representations)
inside an end-to-end module which may incur high variance. In
practice, we adopt a baseline method to reduce the variance of the
predictor which we discuss in Section~\ref{lul}.

\begin{figure}[t]
	\centering
	\scriptsize
	\begin{tabular}{l}
          \hline \\
          utterance:  \textit{which daughter of Barack Obama was named Most Influential Teens in the year 2014} \\
          partially completed logical form: \texttt{and(daughterOf(Barack Obama),  } \\
          next logical form token: \texttt{InfluentialTeensByYear} \\ \\
          \hline \\
          \textbf{soft attention} over all utterance tokens: \\
          \tikz[baseline=(A.base)]{\node[opacity=-1](A) {which};
            \shade[inner color=blue!20] (A.south east) rectangle (A.north west);
            \path (A.center)
            \pgfextra{\pgftext{which}};} 
          \tikz[baseline=(A.base)]{\node[opacity=-1](A) {daughter};
            \shade[inner color=blue!50] (A.south east) rectangle (A.north west);
            \path (A.center)
            \pgfextra{\pgftext{daughter}};} 
          \tikz[baseline=(A.base)]{\node[opacity=-1](A) {of};
            \shade[inner color=blue!30] (A.south east) rectangle (A.north west);
            \path (A.center)
            \pgfextra{\pgftext{of}};} 
          \tikz[baseline=(A.base)]{\node[opacity=-1](A) {Barack};
            \shade[inner color=blue!30] (A.south east) rectangle (A.north west);
            \path (A.center)
            \pgfextra{\pgftext{Barack}};} 
          \tikz[baseline=(A.base)]{\node[opacity=-1](A) {Obama};
            \shade[inner color=blue!50] (A.south east) rectangle (A.north west);
            \path (A.center)
            \pgfextra{\pgftext{Obama}};} 
          \tikz[baseline=(A.base)]{\node[opacity=-1](A) {was};
            \shade[inner color=blue!10] (A.south east) rectangle (A.north west);
            \path (A.center)
            \pgfextra{\pgftext{was}};} 
          \tikz[baseline=(A.base)]{\node[opacity=-1](A) {named};
            \shade[inner color=blue!10] (A.south east) rectangle (A.north west);
            \path (A.center)
            \pgfextra{\pgftext{named}};} 
          \tikz[baseline=(A.base)]{\node[opacity=-1](A) {Most};
            \shade[inner color=blue!10] (A.south east) rectangle (A.north west);
            \path (A.center)
            \pgfextra{\pgftext{Most}};} 
          \tikz[baseline=(A.base)]{\node[opacity=-1](A) {Influential};
            \shade[inner color=blue!60] (A.south east) rectangle (A.north west);
            \path (A.center)
            \pgfextra{\pgftext{Influential}};} 
          \tikz[baseline=(A.base)]{\node[opacity=-1](A) {Teens};
            \shade[inner color=blue!60] (A.south east) rectangle (A.north west);
            \path (A.center)
            \pgfextra{\pgftext{Teens}};} 
          \tikz[baseline=(A.base)]{\node[opacity=-1](A) {in};
            \shade[inner color=blue!10] (A.south east) rectangle (A.north west);
            \path (A.center)
            \pgfextra{\pgftext{in}};} 
          \tikz[baseline=(A.base)]{\node[opacity=-1](A) {the};
            \shade[inner color=blue!20] (A.south east) rectangle (A.north west);
            \path (A.center)
            \pgfextra{\pgftext{the}};} 
          \tikz[baseline=(A.base)]{\node[opacity=-1](A) {Year};
            \shade[inner color=blue!10] (A.south east) rectangle (A.north west);
            \path (A.center)
            \pgfextra{\pgftext{Year}};} 
          \tikz[baseline=(A.base)]{\node[opacity=-1](A) {2014};
            \shade[inner color=blue!30] (A.south east) rectangle (A.north west);
            \path (A.center)
            \pgfextra{\pgftext{2014}};} \\ \\
          \textbf{hard attention} over a single utterance token:\\
          which daughter of Barack Obama was named as the \tikz[baseline=(A.base)]{\node[opacity=-1](A) { Influential };
            \shade[inner color=blue!60] (A.south east) rectangle (A.north west);
            \path (A.center)
            \pgfextra{\pgftext{Influential}};}  Teens in the year 2014
          \\ \\
          \textbf{structured attention} over a subset of utterance tokens:\\
          which  daughter of Barack Obama was named
          \tikz[baseline=(A.base)]{\node[opacity=-1](A) {Most Influential Teens in the year};
            \shade[inner color=blue!60] (A.south east) rectangle (A.north west);
            \path (A.center)
            \pgfextra{\pgftext{Most Influential Teens  in the year}};} 2014
          \\ \\ \hline
	\end{tabular}
	\caption{Different attention mechanisms for predicting the
          next logical form token. The example utterance is
          \textsl{which daughter of Barack Obama was named Most
            Influential Teens in the year 2014?} and the corresponding
          logical form to be generated is
          \texttt{and(daughterOf(Barack Obama),
            InfluentialTeensByYear(2014))}. The figure shows attention
          for predicting \texttt{InfluentialTeensByYear}. Darker
          shading indicates higher values. }
	\label{attvis}
\end{figure}

\paragraph{Binomial Hard Attention}
Learning difficulties aside, a limitation of hard attention lies in
selecting a single token to attend to at each time step. In practice,
a logical form predicate is often triggered by a natural language
phrase or a multi-word expression.  A way to overcome this limitation
is to compute a binomial distribution for every token separately,
indicating the probability of the token being selected.  Then an
attention label is assigned to each token based on this probability
(e.g., with threshold 0.5).  Let $A^i_t \in \{0,1\}$ denote the
attention label of the $i$th token at time step $t$.  Using the
unnormalized attention score $u_t^i$ computed in Equation~(\ref{uti}),
we obtain the probability $p(A_{t}^i = 1)$ as:
\begin{equation}
p(A_{t}^i = 1) = \textnormal{logistic}(u_t^i)
\end{equation}
where \textnormal{logistic} denotes a logistic regression classifier.
We compute adaptive buffer representation as an average of the
selected token embeddings:
\begin{equation}
\bar{b}_t  = \frac{1}{\sum_i A_t^i} \sum_i  A_t^i  b_i 
\end{equation}
which is then used to compute a distribution of the output logical form tokens:
\begin{equation}
y_{t+1}  \sim \textnormal{softmax} (  W_{oy} \tanh( W_f [\bar{b}_t, s_t] )  )
\end{equation}

\subsection{Model Training \label{supervision}}
We now discuss how our neural semantic parser can be trained under
different conditions, i.e.,~with access to utterances annotated
with logical forms, when only denotations are provided, and finally,
when neither logical forms nor denotations are available (see
Table~\ref{training_regime}).
\begin {table}[t]
\caption{Example data for various semantic parsing training regimes.} 
\label{training_regime}
\begin{center}
	\small
	\begin{tabular}{@{}l}
		\hline
		\textbf{Full supervision}: utterance-logical form pairs \\
		utterance: \textit{which daughter of Barack Obama was named Most Influential Teens in the year 2014?} \\
		logical form: \texttt{and(daughterOf(Barack Obama), InfluentialTeensByYear(2014)) } \\
		\\
		\textbf{Weak supervision:} utterance-denotation pairs \\
		utterance: \textit{which daughter of Barack Obama was named Most Influential Teens in the year 2014?}\\
		denotation: \texttt{Malia Obama} \\
		\\
		\textbf{Distant supervision:} entity-masked utterances \\
		utterance: \textit{Malia Obama, the daughter of Barack Obama, was named Most Influential Teens in the year 2014.} \\
		artificial utterance: \textit{\_blank\_, the daughter
          of Barack Obama, was named Most Influential Teens in the year 2014.} \\
		denotation: \texttt{Malia Obama} \\
		\hline
	\end{tabular}
\end{center}
\end{table}

\subsubsection{Learning from Utterance-Logical Form Pairs\label{lul}}
The most straightforward training setup is fully supervised making use
of utterance-logical form pairs.  Consider utterance~$x$ with logical
form~$l$ whose structure is determined by a sequence of transition
operations~$a$ and a sequence of logical form tokens~$y$. Our ultimate
goal is to maximize the conditional likelihood of the logical form
given the utterance for all training data:
\begin{equation}
	\mathcal{L}  = \sum\limits_{(x,l) \in \mathcal{T}}  \log p(l | x) 
\end{equation}
which can be decomposed into the action likelihood and the token likelihood:
\begin{equation}
	\log p(l | x) = \log p(a | x) + \log p(y|x, a)
\end{equation}

\paragraph{Soft attention}
The above objective consists of two terms, one for the action sequence: 
\begin{equation}
	\mathcal{L}_a  = \sum\limits_{(x,l) \in \mathcal{T}}  \log p(a | x) 
	=\sum\limits_{(x,l) \in \mathcal{T}}  \sum\limits_{t=1}^n \log p(a_t | x)
\end{equation}
and one for the logical form token sequence:
\begin{equation}
	\mathcal{L}_y  = \sum\limits_{(x,l) \in \mathcal{T}}  \log p(y | x,a) 
	=\sum\limits_{(x,l) \in \mathcal{T}}  \sum\limits_{t=1}^n \log p(y_t | x,a_t)
\end{equation}
These constitute the training objective for fully differentiable
neural semantic parsers, when (basic or structured) soft attention is
used.

\paragraph{Hard attention}
When hard attention is used for token prediction, the objective
$\mathcal{L}_a$ remains the same but $\mathcal{L}_y$ differs.  This is
because the attention layer is non-differentiable for errors to
backpropagate through.  We use the alternative REINFORCE-style
algorithm \cite{williams1992simple} for backpropagation.  In this
scenario, the neural attention layer is used as a policy predictor to
emit an attention choice, while subsequent neural layers are used as the
value function to compute a reward---a lower bound of the log
likelihood~$\log p(y | x,a)$.  Let~$u_t$ denote the latent attention
choice\footnote{In standard hard attention, the choice is a single token in the sentence; while in binomial hard attention, it is a phrase.} at each time step~$t$; we maximize the expected log likelihood
of the logical form token given the overall attention choice for all
examples, which by Jensen's Inequality is the lower bound on the log
likelihood~$\log p(y | x,a)$:
\begin{equation}
	\begin{split}
		\mathcal{L}_y & =\sum\limits_{(x,l) \in \mathcal{T}}   \sum_{u} \left [ p(u | x,a) \log p(y | u, x,a) \right ] \\
		& \leq \sum\limits_{(x,l) \in \mathcal{T}}  \log \sum_{u} \left [ p(u | x,a) p(y | u, x,a) \right ] \\
		& = \sum\limits_{(x,l) \in \mathcal{T}}  \log p(y | x,a) 
	\end{split}
\end{equation}
The gradient of $\mathcal{L}_y$ with respect to model parameters
$\theta$ is given by:
\begin{equation}
	\begin{split}
		\frac{\partial \mathcal{L}_y}{\partial \theta} & = \sum\limits_{(x,l) \in \mathcal{T}}   \sum_{u}  p(u|x,a) \frac{\partial \log p(y | u, x,a)}{\partial \theta} + \log p(y | u, x,a)\frac{\partial p(u|x,a)}{\partial \theta}   \\ 
		& = \sum\limits_{(x,l) \in \mathcal{T}}   \sum_{u}  p(u|x,a) \frac{\partial \log p(y | u, x,a)}{\partial \theta} + \log p(y | u, x,a)\frac{\partial \log p(u|x,a)}{\partial \theta} p(u|x,a)  \\ 
		& = \sum\limits_{(x,l) \in \mathcal{T}}   \sum_{u} p(u|x,a)  \left [  \frac{\partial \log p(y | u, x,a)}{\partial \theta} +  \log p(y | u, x,a)\frac{\partial \log p(u|x,a)}{\partial \theta} \right ] \\
		& \approx \sum\limits_{(x,l) \in \mathcal{T}}  \frac{1}{N} \sum\limits_{k=1}^K   \left [  \frac{\partial \log p(y | u^k, x,a)}{\partial \theta} +  \log p(y | u^k, x,a)\frac{\partial \log p(u^k|x,a)}{\partial \theta} \right ]
	\end{split}
\end{equation}
which is estimated by the Monte Carlo estimator with $K$ samples.
This gradient estimator incurs high variance because the reward term
$\log p(y|u^k,x,a)$ is dependent on the samples of~$u^k$.  An
input-dependent \textit{baseline} is used to reduce the variance,
which adjusts the gradient update as:
\begin{equation}
	\frac{\partial \mathcal{L}_y}{\partial \theta} =  \sum\limits_{(x,l) \in \mathcal{T}}  \frac{1}{N} \sum\limits_{k=1}^K   \left [  \frac{\partial \log p(y | u^k, x,a)}{\partial \theta} +  (\log p(y | u^k, x,a) - b )\frac{\partial \log p(u^k|x,a)}{\partial \theta} \right ]
\end{equation}
As baseline, we use the soft attention token predictor described
earlier. The effect is to encourage attention samples that return a
higher reward than standard soft attention, while discouraging those
resulting in a lower reward.  For each training case, we approximate
the expected gradient with a single sample of $u^k$.

\subsubsection{Learning from Utterance-Denotation Pairs\label{lud}}
Unfortunately, training data consisting of utterances and their
corresponding logical forms is difficult to obtain at large scale, and
as a result limited to a few domains with a small number of logical
predicates. An alternative to full supervision is a weakly supervised
setting where the semantic parser is trained on utterance-denotation
pairs, where logical forms are treated as latent.

In the following we firstly provide a brief review of conventional
weakly supervised semantic parsing systems \cite{berant2013semantic},
and then explain the extension of our neural semantic parser to a
similar setting.  Conventional weakly-supervised semantic parsing
systems separate the parser from the learner \cite{liang2016learning}.
A chart-based (non-parametrized) parser will recursively build
derivations for each span of an utterance, eventually obtaining a list
of candidate derivations mapping the utterance to its logical form.
The learner (which is often a log-linear model) defines features
useful for scoring and ranking the set of candidate derivations, and
is trained based on the correctness of their denotations.  As
mentioned in \citet{liang2016learning}, the chart-based parser brings
a disadvantage since the system does not support incremental
contextual interpretation, because features of a span can only depend
on the sub-derivations in that span, as a requirement of dynamic
programming.

Different from chart-based parsers, a neural semantic parser is itself
a parametrized model and is able to leverage global utterance
features (via attention) for decoding. However, training the neural
parser directly with utterance-denotation pairs is challenging since
the decoder does not have access to gold standard logical forms for
backpropagation. Moreover, the neural decoder is a conditional
generative model which generates logical forms in a greedy fashion and
therefore lacks the ability to make global judgments of logical forms.  To this end, we
follow conventional setup in integrating our neural semantic parser
with a log-linear ranker, to cope with the weak supervision signal.
The role of the neural parser is to generate a list of candidate
logical forms, while the ranker is able to leverage global features of 
utterance-logical form-denotation triplets
to select which
candidate to use for execution.

The objective of the log-linear ranker is to maximize the log marginal
likelihood of the denotation $d$  via latent logical forms $l$:
\begin{equation}
\log p(d|x) = \log \sum_{l \in L} p(l|x) p(d|x,l) 
\end{equation}
where $L$ denotes the set of candidate logical forms generated by the
neural parser.  Note that $p(d|x,l)$ equates to 1 if the logical form
executes to the correct denotation and 0 otherwise.  For this reason,
we can also write the above equation as
$\log \sum_{l \in L(c)} p(l|x)$, where $L(c)$ is the set of
\textit{consistent} logical forms which execute to the correct
denotation.

Specifically $p(l|x)$ is computed with a
log-linear model:
\begin{equation}
p(l|x) = \frac {\exp (\phi(x, l) \theta )} {\sum_{l' \in L} \exp(\phi (x, l') \theta)}
\end{equation}
where $L$ is the set of candidate logical forms; $\phi$ is the feature function that maps an utterance-logical form pair (and also the corresponding denotation) into a feature vector; and $\theta$ denotes the weight parameter of the model.

Training such a system involves the following steps. Given an input
utterance, the neural parser first generates a list of candidate
logical forms via beam search.  Then these candidate logical forms are
executed and those which yield the correct denotation are marked as
\textit{consistent} logical forms.  The neural parser is then trained
to maximize the likelihood of these consistent logical forms
$\sum_{l \in L_c} \log p(l|x)$.  Meanwhile, the ranker is trained to
maximize the marginal likelihood of denotations $\log p(d|x)$.

Clearly, if the parser does not generate any consistent logical forms,
no model parameters will be updated.  A challenge in this training
paradigm is the fact that we rely exclusively on beam search to find
good logical forms from an exponential search space.  In the beginning
of training, neural parameters are far from optimal, and as a result
good logical forms are likely to fall outside the beam.  We alleviate
this problem by performing entity linking which greatly reduces the
search space. We determine the identity of the entities mentioned in
the utterance according to the knowledge base and restrict the neural
parser to generating logical forms containing only those entities.

\subsubsection{Distant Supervision}
\label{sec:distant-supervision}
Despite allowing to scale semantic parsing to large open-domain
problems
\cite{kwiatkowski2013scaling,berant2013semantic,yao2014information},
the creation of utterance-denotation pairs still relies on
labor-intensive crowd-sourcing. A promising research direction is to employ 
a sort of distant supervision, where training data (e.g., artificial utterance-denotations pairs)
is artificially generated with give resources (e.g., a knowledge base, Wikipedia documents).
In this work, we additionally train the weakly-supervised neural semantic parser 
with a distant supervision approach proposed by \citet{reddy2014large}. 
In this setting, the
given data is a corpus of
entity-recognized sentences and a knowledge
base.  Utterance-denotation pairs are artificially created by
replacing entity mentions in the sentences with variables.  Then, the
semantic parser is trained to predict the denotation for the variable
that includes the mentioned entity. For example,
given the declarative sentence \textsl{NVIDIA was founded by Jen-Hsun
	Huang and Chris Malachowsky}, the distant supervision approach
creates the utterance \textsl{NVIDIA was founded by Jen-Hsun\_Huang and
	\_blank\_} paired with the corresponding denotation \textsl{Chris
	Malachowsky}. 
In some cases, even stronger
constraints can be applied. For example, if the mention is preceded by
the word \textsl{the}, then the correct denotation includes exactly
one entity. In sum, the approach converts the corpus of
entity-recognized sentences into artificial utterance-denotation pairs on which
the weakly supervised model described in Section~\ref{lud} can be
trained. We also aim to evaluate if this approach is helpful for 
practical question answering.


\section{Experiments}
\label{sec:experiments}

In this section, we present our experimental setup for assessing the
performance of the neural semantic parsing framework. We
present the datasets on which our model was trained and tested,
discuss implementation details,   and finally report and analyze
semantic parsing results.

\subsection{Datasets}
\label{sec:datasets}
We evaluated our model on the following datasets which cover different
domains and require different types of supervision.

\textsc{GeoQuery} \cite{zelle1996learning} contains 880 questions and
database queries about US geography. The utterances are compositional,
but the language is simple and vocabulary size small (698~entities and
24~relations). Model training on this dataset is fully supervised
(Section~\ref{lul})

\textsc{WebQuestions} \cite{berant-EtAl:2013:EMNLP} contains~5,810
question-answer pairs.  It is based on Freebase and the questions are
not very compositional. However, they are real questions asked by
people on the web.  

\textsc{GraphQuestions} \cite{su2016generating} contains 5,166
question-answer pairs which were created by showing 500~Freebase graph
queries to Amazon Mechanical Turk workers and asking them to
paraphrase them into natural language. Model training on
\textsc{WebQuestions} and \textsc{GraphQuestions} is weakly supervised
(Section~\ref{lud}).

\textsc{Spades} \cite{bisk2016evaluating} contains 93,319 questions
derived from \textsc{clueweb09} \cite{gabrilovich2013facc1}
sentences. Specifically, the questions were created by randomly
removing an entity, thus producing sentence-denotation pairs
\cite{reddy2014large}.  The sentences include two or more entities and
although they are not very compositional, they constitute a
large-scale dataset for neural network training with distant
supervision (Section~\ref{sec:distant-supervision}).

\subsection{Implementation Details}
\label{sec:impl-deta}

\paragraph{Shared Parameters}
Across training regimes, the dimensions of word vector, logical form
token vector, and LSTM hidden state are 50, 50, and~150 respectively.
Word embeddings were initialized with Glove embeddings
\cite{pennington2014glove}.  All other embeddings were randomly
initialized.  We used one LSTM layer in forward and backward
directions.  Dropout was used on the combined feature representation
of the buffer and the stack (Equation (\ref{softmax})), which computes
the softmax activation of the next action or token.  The dropout rate
was set to 0.5.  Finally, momentum SGD \cite{sutskever2013importance}
was used as the optimization method to update the parameters of the
model.

\paragraph{Entity Resolution}
Amongst the four datasets described above, only \textsc{GeoQuery}
contains annotated logical forms which can be used to directly train a
neural semantic parser.  For the other three datasets, supervision is
indirect via consistent logical forms validated on denotations (see
Section~\ref{lud}). As mentioned earlier, we use entity linking to
reduce the
search space for consistent logical forms.   Entity mentions in \textsc{Spades} are
automatically annotated with Freebase entities
\cite{gabrilovich2013facc1}. For \textsc{WebQuestions} and
\textsc{GraphQuestions} we perform entity linking following the
procedure described in \citet{reddy2016transforming}. We identify
potential entity spans using seven handcrafted part-of-speech patterns
and associate them with Freebase entities obtained from the
Freebase/KG API (\url{http://developers.google.com/freebase/}).  For
each candidate entity span, we retrieve the top 10 entities according
to the API.  We treat each possibility as a candidate entity to
construct candidate utterances with beam search of size 500, among
which we look for the consistent logical forms.

\paragraph{Discriminative Ranker}
For datasets which use denotations as  supervision, our semantic
parsing system additionally includes a discriminative ranker, whose
role is to select the final logical form to execute from a list of
candidates generated by the neural semantic parser. At test time, the
generation process is accomplished by beam search with beam size 300.
The ranker which is a log-linear model is trained with momentum SGD
\cite{sutskever2013importance}.  As features, we consider the
embedding cosine similarity between the utterance (excluding
stop-words) and the logical form, the token overlap count between the
two, and also similar features between the lemmatized utterance and
the logical form.  In addition, we include as features the embedding
cosine similarity between the question words and the logical form, the
similarity between the question words (e.g., \textsl{what, who, where,
  whose, date, which, how many, count}) and relations in the logical
form, and the similarity between the question words and answer type as
indicated by the last word in the Freebase relation
\cite{xu2016question}.  Finally, we add as a feature the length of the
denotation given by the logical form \cite{berant2013semantic}.

\subsection{Results} 
\label{sec:results}

In this section, we present the experimental results of our
\textbf{T}ransition-based \textbf{N}eural \textbf{S}emantic
\textbf{P}arser (\textsc{tnsp}). We present various instantiations of
our own model as well as comparisons against semantic parsers proposed
in the literature.


\begin {table}[t]
\begin{center}
  \caption{Fully supervised experimental results on the
    \textsc{GeoQuery} dataset. For \citet{jia2016data}, we include two
    of their results: one is a standard neural sequence to sequence
    model; and the other is the same model trained with a data
    augmentation algorithm on the labeled data (reported in
    parentheses).}
\label{geo}
	\begin{tabular}{lc}
		\hline
		\multicolumn{1}{c}{Models} & \multicolumn{1}{c}{Accuracy} \\ \hline \hline
		\citet{zettlemoyer:learning:2005} & 79.3 \\ 
		\citet{zettlemoyer2007online} & 86.1 \\
		\citet{kwiatkowksi2010inducing} & 87.9\\
		\citet{kwiatkowski2011lexical} & 88.6\\
		\citet{kwiatkowski2013scaling} & 88.0\\
		\citet{zhao2014type} & 88.9 \\
		\citet{liang2011learning} & 91.1 \\\hline \hline
		\citet{dong2016language}  & 84.6 \\
		\citet{jia2016data} & \hspace*{5.6ex}85.0 (89.1)\\
		\citet{Rabinovich-Stern-Klein:2017:AbstractSyntaxNetworks}
                & 87.1 \\ \hline \hline
		\textsc{tnsp}, soft attention, top-down & 86.8 \\
		\textsc{tnsp}, soft structured attention, top-down & 87.1 \\
		\textsc{tnsp}, hard attention, top-down & 85.3 \\
		\textsc{tnsp}, binomial hard attention, top-down & 85.5 \\
		\textsc{tnsp}, soft attention, bottom-up & 86.1 \\
		\textsc{tnsp}, soft structured attention, bottom-up & 86.8 \\
		\textsc{tnsp}, hard attention, bottom-up & 85.3 \\
		\textsc{tnsp}, binomial hard attention, bottom-up & 85.3 \\
		\hline
	\end{tabular}
\end{center}
\end{table}

Experimental results on \textsc{GeoQuery} are shown in
Table~\ref{geo}.  The first block contains conventional statistical
semantic parsers, previously proposed neural models are presented in
the second block, whereas variants of \textsc{tnsp} are shown in the
third block.  Specifically we build various top-down and bottom-up
\textsc{tnsp} models using the various types of attention introduced
in Section~\ref{sec:next-token-pred}. We report accuracy which is
defined as the proportion of utterances which correctly parsed to
their gold standard logical forms. Amongst \textsc{tnsp} models, a
top-down system with structured (soft) attention performs
best. Overall, we observe that differences between top-down and
bottom-up systems are small; it is mostly the attention mechanism that
affects performance, with hard attention performing worst and soft
attention performing best for both top-down and bottom-up
systems. \textsc{tnsp} outperforms previously proposed neural semantic
parsers which treat semantic parsing as a sequence transduction
problem and use LSTMs to map utterances to logical forms
\cite{dong2016language,jia2016data}.  \textsc{tnsp} brings performance
improvements over these systems when using comparable data sources for
training. \citet{jia2016data} achieve better results with synthetic
data that expands \textsc{GeoQuery}; we could adopt their approach to
improve model performance, however, we leave this to future work.  Our
system is on the same par with the model of
\citet{Rabinovich-Stern-Klein:2017:AbstractSyntaxNetworks} who also
output well-formed trees in a top-down manner using a decoder built of
many submodels, each associated with a specific construct in the
underlying grammar.

Results for the weakly supervised training scenario are shown in
Table~\ref{weaksup}. For all Freebase related datasets we use average
F1 \cite{berant2013semantic} as our evaluation metric.  We report
results on \textsc{WebQuestions} and \textsc{GraphQuestions} in
Tables~\ref{webqa} and~\ref{graphqa}, respectively. The first block in
the tables groups conventional statistical semantic parsers, the second
block presents related neural models, and the third block variants of
\textsc{tnsp}. For fair comparison, we also built a baseline
sequence-to-sequence model enhanced with an attention mechanism
\cite{dong2016language}. 

On \textsc{WebQuestions}, the best performing \textsc{tnsp} system
generates logical forms based on top-down pre-order while employing
soft attention. The same top-down system with structured attention
performs closely. Again we observe that bottom-up preorder lags
behind. In general, our semantic parser obtains performance on par
with the best symbolic systems (see the first block in
Table~\ref{webqa}). It is important to note that \citet{bast2015more}
develop a question answering system, which contrary to ours cannot
produce meaning representations whereas \citet{berant2015imitation}
propose a sophisticated agenda-based parser which is trained borrowing
ideas from imitation learning.  \citet{reddy2016transforming} learn a
semantic parser via intermediate representations which they generate
based on the output of a dependency parser. \textsc{tnsp} performs
competitively despite not having access to linguistically-informed
syntactic structure. Regarding neural systems (see the second block in
Table~\ref{webqa}), our model outperforms the sequence-to-sequence
baseline and other related neural architectures using similar
resources.  \citet{xu2016question} represent the state of the art on
\textsc{webquestions}. Their system uses Wikipedia to prune out
erroneous candidate answers extracted from Freebase. Our model would
also benefit from a similar post-processing.

\begin {table}[t]
\begin{center}
\captionsetup{width=\textwidth}
\caption{Weakly supervised experimental results on two
  datasets. Results with additional resources are shown in
  parentheses. \label{weaksup}}
\small \subfloat[ \textsc{WebQuestions} \label{webqa}]{
	\begin{tabular}{lc}\hline
 	\multicolumn{1}{c}{Models} & \multicolumn{1}{c}{F1} \\ \hline\hline
	 	\textsc{Sempre} \cite{berant-EtAl:2013:EMNLP} & 35.7\\
	 	\textsc{Jacana} \cite{yao2014information} & 33.0 \\
	 	\textsc{ParaSempre}\cite{berant2014semantic} & 39.9 \\
	 	\textsc{Aqqu} \cite{bast2015more} & 49.4 \\
	 	\textsc{AgendaIL} \cite{berant2015imitation} & 49.7 \\
	 	\textsc{DepLambda} \cite{reddy2016transforming} & 50.3 \\\hline \hline
	 	\textsc{SubGraph} \cite{bordesquestion} & 39.2 \\
	 	\textsc{mccnn} \cite{dong2015question} & 40.8 \\
	 	\textsc{stagg} \cite{yih2015semantic} & 52.5 \\
	 	\textsc{mcnn} \cite{xu2016question} &
                53.3 \\ 
          Sequence-to-sequence & 48.3 \\\hline \hline
          \textsc{tnsp}, soft attention, top-down & 50.1 \\
          \textsc{tnsp}, soft structured attention, top-down & 49.8 \\
          \textsc{tnsp}, hard attention, top-down & 49.4 \\
          \textsc{tnsp}, binomial hard attention, top-down & 48.7 \\
          \textsc{tnsp}, soft attention, bottom-up &  49.6\\
          \textsc{tnsp}, soft structured attention, bottom-up & 49.5\\ 
          \textsc{tnsp}, hard attention, bottom-up & 48.4\\
          \textsc{tnsp}, binomial hard attention, bottom-up & 48.7\\\hline

\end{tabular}
} \\
\subfloat[ \textsc{GraphQuestions} \label{graphqa} ]{
	\begin{tabular}{lr}\hline
          \multicolumn{1}{c}{Models} & \multicolumn{1}{c}{F1} \\ \hline\hline
          \textsc{sempre} \cite{berant-EtAl:2013:EMNLP} & 10.8 \\ %
          \textsc{parasempre} \cite{berant2014semantic} & 12.8 \\
          \textsc{jacana} \cite{yao2014information} & 5.1 \\
          \textsc{SimpleGraph} \cite{reddy2016transforming}  & 15.9 \\ 
          \textsc{UDepLambda} \cite{reddy2017universal} & 17.6 \\\hline\hline
          Sequence-to-sequence & 16.2 \\
          \textsc{Para4QA} \cite{dong-EtAl:2017:EMNLP2017} & 20.4 \\ \hline \hline
          \textsc{tnsp}, soft attention, top-down & 17.3 \\
          \textsc{tnsp}, soft structured attention, top-down & 17.1 \\ 
          \textsc{tnsp}, hard attention, top-down & 16.2\\
          \textsc{tnsp}, binomial hard attention, top-down & 16.4\\
          \textsc{tnsp}, soft attention, bottom-up & 16.9 \\
          \textsc{tnsp}, soft structured attention, bottom-up & 17.1 \\ 
          \textsc{tnsp}, hard attention, bottom-up & 16.8\\
          \textsc{tnsp}, binomial hard attention, bottom-up & 16.5\\
          \hline
	\end{tabular}
}
\end{center}
\end{table}

With respect to \textsc{GraphQuestions}, we report F1~for various
\textsc{tnsp} models (third block in Table~\ref{graphqa}), and conventional statistical semantic parsers (first block in Table~\ref{graphqa}). The first three
systems are presented in \citet{su2016generating}. Again, we
observe that a top-down variant of \textsc{tnsp} with soft attention
performs best. It is superior to the sequence-to-sequence baseline and
obtains performance comparable to \citet{reddy2017universal} without
making use of an external syntactic parser. The model of
\citet{dong-EtAl:2017:EMNLP2017} is state of the art on
\textsc{GraphQuestions}. Their method is trained end-to-end using
questions-answer pairs as a supervision signal together with question
paraphrases as a means of capturing different ways of expressing the
same content. Importantly, their system is optimized with
question-answering in mind, and does not produce logical forms.

\begin{figure}[t]
	\center
	\begin{tikzpicture}
	\begin{axis}[ymin=0,ymax=1, xmin=0, xmax=500, legend cell align={left}, xlabel={beam size}, ylabel={fraction}]
	\addplot[color= blue,mark=*,nodes near coords,
	every node near coord/.append style={
		/pgf/number format/fixed zerofill,
		/pgf/number format/precision=3
	}
	] coordinates { (0,0) (50,401/764.0) (100,439/764.0) (200,463/764.0) (300,486/764.0) (400,493/764.0) (500,503/764.0)};
	\addlegendentry{answerable}
	\addplot[color= red,mark=square*,nodes near coords,
	every node near coord/.append style={
		/pgf/number format/fixed zerofill,
		/pgf/number format/precision=3
	}
	] coordinates { (0,0) (50,116/764.0) (100,149/764.0) (200,163/764.0) (300,170/764.0) (400,173/764.0) (500,173/764.0) };
	\addlegendentry{correctly answered}
	\end{axis}
	\end{tikzpicture}
	\caption{Fraction of utterances that are answerable versus
          those correctly predicted with varying beam size on the
          \textsc{GraphQuestions} development set.}
	\label{beam}
\end{figure}
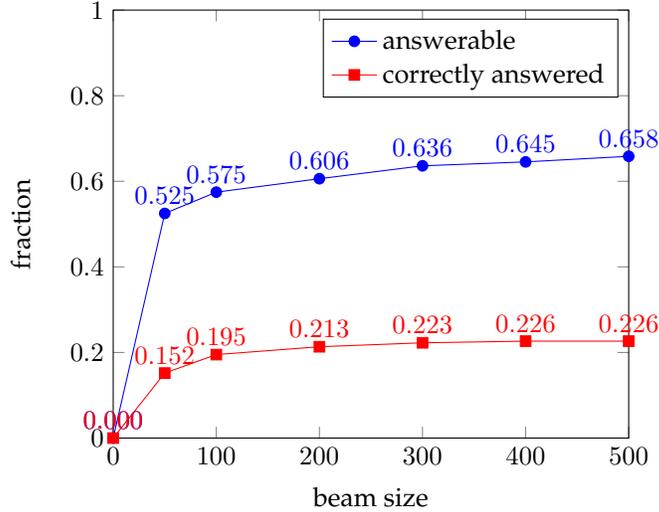

When learning from denotations, a challenge concerns the handling of an
exponentially large set of logical forms.  In our approach, we rely on
the neural semantic parser to generate a list of candidate logical
forms by beam search.  Ideally, we hope the beam size is large enough
to include good logical forms which will be subsequently selected by
the discriminative ranker.  Figure~\ref{beam} shows the effect of
varying beam size on \textsc{GraphQuestions} (development set) when
training executes for two epochs using the \textsc{tnsp} soft
attention model with top-down generation order.  We report the number
of utterances that are answerable (i.e., an utterance is considered
answerable if the beam includes one or more good logical forms leading
to the correct denotation) and the number of utterances that are
correctly answered eventually. As the beam size increases, the gap
between utterances that are answerable and those that are answered
correctly becomes larger. And the curve for correctly answered
utterances gradually plateaus and the performance does not improve.
This indicates a trade-off between generating candidates that cover
good logical forms and picking the best logical form for execution:
when the beam size is large, there is a higher chance for good logical
forms to be included but also for the discriminative ranker to make
mistakes.

\begin {table}[t]
\begin{center}
  \caption{Breakdown of questions answered by type for the
    \textsc{GraphQuestions}.}
\label{answer by type}
	\begin{tabular}{l r c c}
		\hline 
		Question type & Number & \% Answerable & \% Correctly answered \\
		\texttt{relation} & 1938 &  0.499 & 0.213 \\
		\texttt{count} & 309 & 0.421 & 0.032 \\
		\texttt{aggregation} & 226 &  0.363 & 0.075 \\
		\texttt{filter} &135 & 0.459 & 0.096 \\
		All & 2,608 & 0.476 & 0.173 \\
		\hline
	\end{tabular}
\end{center}
\end{table}

\textsc{GraphQuestions} consists of four types of questions.  As shown
in Table~\ref{answer by type}, the first type are relational questions
(denoted by \texttt{relation}). An example of a relational question is
\textsl{what periodic table block contains oxygen}; the second type
contains count questions (denoted by \texttt{count}). An example is
\textsl{how many firefighters does the new york city fire department
  have}; the third type includes aggregation questions requiring
argmax or argmin (denoted by \texttt{aggregation}). An example is
\textsl{what human stampede injured the most people}; the last type
are filter questions which requires comparisons by $>$, $\geq$, $<$
and $\leq$ (denoted by \texttt{filter}). An example is \textsl{which
  presidents of the united states weigh not less than 80.0 kg}.
Table~\ref{answer by type} shows the number of questions broken down
by type, as well as the proportion of answerable and correctly
answered questions.  As the results reveal, \texttt{relation}
questions are the simplest to answer which is expected since
\texttt{relation} questions are non-compositional and their logical
forms are easy to find by beam search.  The remaining types of
questions are rather difficult to answer: although the system is able
to discover logical forms that lead to the correct denotation during
beam search, the ranker is not able to identify the right logical
forms to execute. Aside from the compositional nature of these
questions which makes them hard to answer, another difficulty is that
such questions are a minority in the dataset posing a learning
challenge for the ranker to identify them. As future work, we plan to
train separate rankers for different question types.

\begin {table}[t]
\caption{Distantly supervised experimental results on the \textsc{Spades} dataset.}
\label{spade}
\begin{center}
  \begin{tabular}{lc}
		\hline 
		\multicolumn{1}{c}{Models} & \multicolumn{1}{c}{F1} \\ \hline\hline
		Unsupervised CCG \cite{bisk2016evaluating} & 24.8 \\
		Semi-supervised CCG \cite{bisk2016evaluating} & 28.4 \\
		Supervised CCG \cite{bisk2016evaluating} & 30.9 \\
		Rule-based system \cite{bisk2016evaluating} & 31.4
                \\\hline \hline
		Sequence-to-sequence & 28.6 \\ \hline \hline
		\textsc{tnsp}, soft attention, top-down & 32.4 \\
		\textsc{tnsp}, soft structured attention, top-down & 32.1 \\
		\textsc{tnsp}, hard attention, top-down & 31.5 \\
		\textsc{tnsp}, binomial hard attention, top-down & 29.8 \\
		\textsc{tnsp}, soft attention, bottom-up &   32.1 \\
		\textsc{tnsp}, soft structured attention, bottom-up & 31.4\\ 
		\textsc{tnsp}, hard attention, bottom-up & 30.7 \\
		\textsc{tnsp}, binomial hard attention, bottom-up & 30.4
		\\ \hline
	\end{tabular}
\end{center}
\end{table}

Finally, Table~\ref{spade} presents experimental results on
\textsc{Spades} which serves as a testbed for our distant supervision
setting. Previous work on this dataset has used a semantic parsing
framework where natural language is converted to an intermediate
syntactic representation and then grounded to Freebase. Specifically,
\citet{bisk2016evaluating} evaluate the effectiveness of four
different CCG parsers on the semantic parsing task when varying the
amount of supervision required. As can be seen, \textsc{tnsp}
outperforms all CCG variants (from unsupervised to fully supervised)
without having access to any manually annotated derivations or
lexicons. Again, we observe that a top-down \textsc{tnsp} system with
soft attention performs best and is superior to the
sequence-to-sequence baseline.

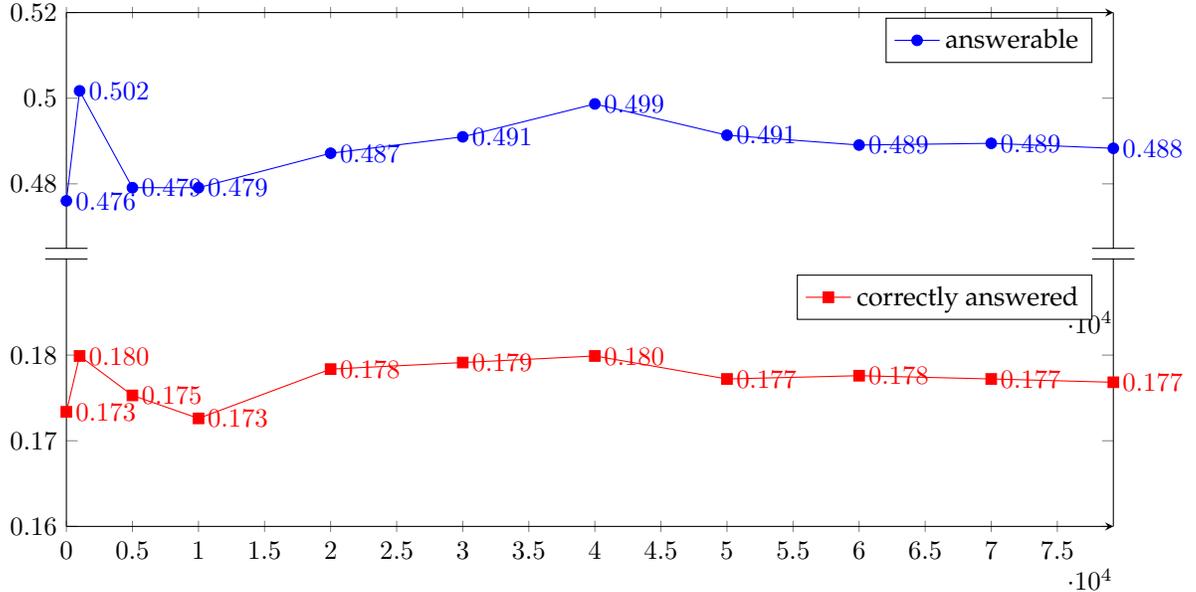
\begin{figure}[t]
	\centering
\begin{tikzpicture}
\begin{groupplot}[
group style={
	group name=my fancy plots,
	group size=1 by 2,
	xticklabels at=edge bottom,
	vertical sep=0pt
},
width=15.5cm,
xmin=0
]

\nextgroupplot[ymin=0.46,ymax=0.52,
ytick={0.48,0.5,0.52},
axis x line=top, 
axis y discontinuity=parallel,
height=5cm]
\addplot[color= blue,mark=*,nodes near coords,
nodes near coords align=horizontal,
every node near coord/.append style={
	/pgf/number format/fixed zerofill,
	/pgf/number format/precision=3
}
] coordinates { (0,1241/2607.0) (1000,1308/2607.0) (5000,1249/2607.0) (10000,1249/2607.0) (20000,1270/2607.0) (30000,1280/2607.0) (40000,1300/2607.0) (50000,1281/2607.0) (60000,1275/2607.0) (70000,1276/2607.0) (79243,1273/2607.0)};       
\addlegendentry{answerable}

\nextgroupplot[ymin=0.16,ymax=0.19,
ytick={0.16, 0.17, 0.18},
axis x line=bottom,
height=5cm]
\addplot[color= red,mark=square*,nodes near coords,
nodes near coords align=horizontal,
every node near coord/.append style={
	/pgf/number format/fixed zerofill,
	/pgf/number format/precision=3
}
] coordinates { (0,452/2607.0) (1000,469/2607.0) (5000,457/2607.0) (10000,450/2607.0) (20000,465/2607.0) (30000,467/2607.0) (40000,469/2607.0) (50000,462/2607.0) (60000,463/2607.0) (70000,462/2607.0) (79243,461/2607.0)};          
\addlegendentry{correctly answered}
\end{groupplot}
\end{tikzpicture}
\caption{Fraction of answerable and correctly answered questions in the
	\textsc{GraphQuestions} when different amount of the
	\textsc{Spades} data is used.}
\label{supervision test}
\end{figure}

The results on \textsc{Spades} hold promise for scaling semantic
parsing by using distant supervision.  In fact, artificial data could potentially
help improve weakly supervised question answering models trained on
utterance-denotation pairs. To this end, we use the entity-masked
declarative sentences paired with their denotations in \textsc{Spades}
as additional training data for \textsc{GraphQuestions}.  We train the
neural semantic parser with the combined training data and evaluate on
the \textsc{GraphQuestions}.  We use the top-down, soft-attention
\textsc{TNSP} model with a beam search size of~300.  During each epoch
of training, the model was first trained with a mixture of the
additional \textsc{Spades} data and the original training data.
Figure~\ref{supervision test} shows the fraction of answerable and
correctly answered questions generated by the neural semantic parser
on \textsc{GraphQuestions}.  Note that the original
\textsc{GraphQuestions} training set consists of 1,794 examples and we
report numbers when different amount of \textsc{Spades} training data
is used.

As the figure shows, using artificially training data is able to
improve the neural semantic parser on a question answering task to
some extent.  This suggests that distant supervision is a promising
direction for building practical semantic parsing systems.  Since
artificial training data can be abundantly generated to fit a neural
parser, the approach can be used for data argumentation when
question-answer pairs are limited. 

However, we observe that the
maximum gain occurs when 1,000~extra training examples are used, a
size comparable to the original training set.  After that no further
improvements are made when more training examples are used.  We
hypothesize this is due to the disparities between
utterance-denotation pairs created in distant supervision and
utterance-denotation pairs gathered from real users.  For example,
given the declarative sentence \textsl{NVIDIA was founded by Jen-Hsun
  Huang and Chris Malachowsky}, the distant supervision approach
creates the utterance \textsl{NVIDIA was founded by Jen-Hsun\_Huang
  and \_blank\_} and the corresponding denotation \textsl{Chris
  Malachowsky}.  However, the actual question users may ask is
\textsl{Who founded NVIDIA together with Jen-Hsun\_Huang}.  This poses
a challenge if the neural network is trained on one type of utterance
and tested on another. We observe that the distribution mismatch
outweighs the addition of artificial data quickly.  Future work will
focus on how to alleviate this problem by generating more realistic
data with an advanced question generation module.  

Another factor
limiting performance is that \textsc{Spades} mainly consists of
relational questions without high-level predicates, such as
\texttt{count}, \texttt{filter} and \texttt{aggregation} which
substantially harder to answer correctly (see Table~\ref{answer by
  type}).


To summarize, across experiments and training regimes, we observe that
\textsc{tnsp} performs competitively while producing meaningful and
well-formed logical forms.  One characteristic of the neural semantic
parser is that it generates tree-structured representations in an
arbitrarily canonical order, as a sequence of transition
operations. We investigated two such orders, top-down pre-order and
bottom-up post-order.  Experimentally, we observed that pre-order
generation provides marginal benefits over post-order generation. One
reason for this is that compared to sibling information which the
bottom-up system uses, parent information used by the top-down system
is more important for subtree prediction.

\begin {table}[t]
\begin{center}
	\small
	\begin{tabular}{l}
		
		\textbf{hard attention}  \\
		\hline
		good selections:  \\ 
		\hline
		the brickyard 400 was hosted at what \textcolor{blue}{\textbf{venue}}? (\texttt{base.nascar.nascar\_venue})\\
		christian faith branched from what \textcolor{blue}{\textbf{religion}}?  (\texttt{religion.religion})\\
		which \textcolor{blue}{\textbf{paintings}} are discovered in lascaux? (\texttt{base.caveart.painting})\\
		\hline
		bad selections: \\
		\hline
		which \textcolor{red}{\textbf{violent}} events started on 1995-04-07? (\texttt{base.disaster2.attack})\\
		who was \textcolor{red}{\textbf{the}} aircraft designer of the b-747? (\texttt{aviation.aircraft\_designer})\\
		the boinc has been \textcolor{red}{\textbf{used}} in which services? (\texttt{base.centreforeresearch.service})\\
		\hline
		neutral selections: \\
		\hline
		how does ultram \textbf{act} in the body? (\texttt{medicine.drug\_mechanism\_of\_action})\\
		microsoft has created which \textbf{programming} languages? (\texttt{computer.programming\_language})\\
		find un agencies \textbf{founded} in 1957 (\texttt{base.unitednations.united\_nations\_agency}).
		\\ \\
		\textbf{structured attention} \\
		\hline
		good selections:  \\ 
		\hline
		the brickyard 400 was \textcolor{blue}{\textbf{hosted at what venue}}? (\texttt{base.nascar.nascar\_venue})\\
		which \textcolor{blue}{\textbf{violent events started on}} 1995-04-07?  (\texttt{base.disaster2.attack})\\
		how does ultram \textcolor{blue}{\textbf{act in the body}}? (\texttt{medicine.drug\_mechanism\_of\_action})\\
		\hline
		bad selections: \\
		\hline
		\textcolor{red}{\textbf{what is}} ehrlich's \textcolor{red}{\textbf{affiliation}}? (\texttt{education.department})\\
		\textcolor{red}{\textbf{for which war was the}} italian armistice \textcolor{red}{\textbf{signed}}? (\texttt{base.morelaw.war})\\
		the boinc \textcolor{red}{\textbf{has been used in}} which services? (\texttt{base.centreforeresearch.service})\\
		\hline
		neutral selections: \\
		\hline
		\textbf{where was the} brickyard 400 \textbf{held}? (\texttt{base.nascar.nascar\_venue})\\
		\textbf{by whom was} paul \textbf{influenced}? (\texttt{influence.influence\_node})\\
		\textbf{how does} ultram \textbf{act in the body}? (\texttt{medicine.drug\_mechanism\_of\_action})
		
	\end{tabular}
\end{center}
\caption{Hard attention and structure attention when predicting the
  relation in each question. The corresponding logical predicate is shown in brackets.}
\label{attchoice}
\end{table}

We explored three attention mechanisms in our work, including soft
attention, hard attention, and structured attention.  Quantitatively,
we observe that soft attention always outperforms hard attention in
all three training setups. This can be attributed to the
differentiability of the soft attention layer. The structured
attention layer is also differentiable since it computes the marginal
probability of each token being selected with a dynamic programming
procedure. We observe that on \textsc{GeoQuery} which represents the
fully supervised setting, structured attention offers marginal gains
over soft attention. But in other datasets where logical forms are not
given, the more structurally aware attention mechanism does not
improve over soft attention, possibly due to the weaker supervision
signal. However, it should be noted that the structured attention
layer at each decoding step requires the forward-backward algorithm,
which has time complexity $O(2n^2)$ (where~$n$ denotes the utterance
length) and therefore much slower than soft attention which has linear
($O(n)$) complexity.

An advantage of hard and structured attention is that it allows us to
inspect which natural language tokens are being selected when
predicting a relation or entity in the logical form.  For hard
attention, the selection boils down to a token sampling procedure;
whereas for structured attention, the tokens selected can be
interpreted with the Viterbi algorithm which assigns the most likely
label for each token.  Table~\ref{attchoice} shows examples of hard
and structured attention when predicting the key relational logical
predicate. These examples were selected from \textsc{GraphQuestions}
using the top-down \textsc{tnsp} system. The table contains both
meaningful token selections (where the selected tokens denote an
informative relation) and non-meaningful ones.




\section{Conclusions}
\label{sec:disc-concl}

In this paper, we described a general neural semantic parsing
framework which operates with functional query language and generates
tree-structured logical forms with transition-based neural networks.
To tackle mismatches between natural language and logical form tokens,
we introduced various attention mechanisms in the generation process.
We also considered different training regimes, including fully
supervised training where annotated logical forms are given,
weakly-supervised training when denotations are provided, and distant
supervision where only unlabeled sentences and a knowledge base are
available. Compared to previous neural semantic parsers, our model
generates well-formed logical forms, and is more interpretable ---
hard and structured attention can be used to inspect what the model
has learned.

When the training data consists of utterance-denotation pairs, we
employ a generative parser-discriminative ranker framework: the role
of the parser is to (beam) search for candidate logical forms, which
are subsequently re-scored by the ranker. This is in contrast to
recent work \cite{neelakantan2017learning} on weakly-supervised neural
semantic parsing, where the parser is directly trained by
reinforcement learning using denotations as reward. Advantageously,
our framework employs beam search (in contrast to greedy decoding) to
increase the likelihood of discovering correct logical forms in a
candidate set. Meanwhile, the discriminative ranker is able to
leverage \textit{global} features on utterance-logical form-denotation
triplets to score logical forms.  In future, we will compare the
presented parser-ranker framework with reinforcement learning-based
parsers.

Directions for future work are many and varied. Since the current
semantic parser generates tree structured logical forms conditioned on
an input utterance, we could additionally exploit input information
beyond sequences such as dependency tree representations, resembling a
tree-to-tree transduction model.  To tackle long-term dependencies in
the generation process, an intra-attention mechanism could be used
\cite{cheng2016long,vaswani2017attention}.  Secondly, when learning
from denotations, it is possible that the beam search output contains
spurious logical forms which lead to correct answers accidentally but
do not represent the actual meaning of an utterance. Such logical
forms are misleading training signals and should be removed,
e.g.,~with a generative neural network component
\cite{cheng2017generative} which scores how well a logical form
represents the utterance semantics. 
 Last but not least, since our
semantic parsing framework provides a decomposition between
domain-generic tree generation and the selection of domain-specific
constants, we would like to further explore training the semantic
parser in a muti-domain setup \cite{herzig2017neural}, where the
domain-generic parameters are shared.

\starttwocolumn
\bibliography{compling_style}

\end{document}